\documentclass[letterpaper, 10 pt, conference]{ieeeconf}
\IEEEoverridecommandlockouts                              
\usepackage[utf8]{inputenc}
\let\labelindent\relax
\usepackage{amsfonts}       

\usepackage{amsthm}
\usepackage{mathtools}      
\usepackage{amssymb}        
\usepackage{filecontents}
\usepackage{graphicx}       
\usepackage{marginnote}     
\usepackage{marvosym}       
\usepackage{overpic}        
\usepackage{tabularx}
\usepackage{subfigure}
\usepackage{cite}
\usepackage{color}
\usepackage[normalem]{ulem}
\usepackage{epstopdf}
\usepackage{enumitem}
\usepackage[normalem]{ulem}
\usepackage{lipsum}
\usepackage{url}
\usepackage[hidelinks]{hyperref}
\usepackage[dvipsnames]{xcolor}
\usepackage{diagbox}
\usepackage{multirow}
\usepackage{microtype}
\usepackage{xspace}
\usepackage{comment}

\makeatletter
\newif\if@restonecol
\makeatother

\usepackage[linesnumbered,ruled,vlined]{algorithm2e}
\usepackage{algpseudocode}
\usepackage{amsmath}
\usepackage{enumitem}
\setlist[itemize]{itemsep=1pt,topsep=1pt}
\setlist[enumerate]{itemsep=1pt,topsep=1p}

\newif\ifdraft
\draftfalse

\ifdraft
\usepackage[paperheight=11in,paperwidth=9.5in,
			left=1.25in,right=1.25in,
			top=0.75in,bottom=0.75in,
			heightrounded,marginparwidth=1.2in,
			marginparsep=0.05in]{geometry}
\usepackage{xcolor}
\usepackage{xargs} 
\usepackage[textsize=footnotesize]{todonotes}
\newcommandx{\sh}[2][1=]{\todo[linecolor=blue,
			backgroundcolor=blue!10,bordercolor=blue,#1]{Han: #2}}
\newcommandx{\tg}[2][1=]{\todo[linecolor=orange,
			backgroundcolor=orange!10,bordercolor=orange,#1]{Greaten: #2}}
\newcommandx{\jy}[2][1=]{\todo[linecolor=green,
			backgroundcolor=green!10,bordercolor=green,#1]{JJ: #2}}
\else
\newcommand{\sh}[1]{{}}
\newcommand{\tg}[1]{{}}
\newcommand{\jy}[1]{{}}
\fi

\newif\iftwocolumn
\twocolumntrue

\newtheorem{problem}{Problem}
\newtheorem{proposition}{Proposition}[section]
\newtheorem{lemma}{Lemma}[section]

\newtheorem*{remark}{Remark}

\makeatletter
\def\subsubsection{\@startsection{subsubsection}
                                 {3}
                                 {\z@ \hspace*{1mm}}
                                 {0ex plus 0.1ex minus 0.1ex}
                                 {0ex}
                                 {\normalfont\normalsize\itshape}}
\makeatother

\newcommand{\mpp}{\textsc{MRMP}\xspace}
\newcommand{\ilp}{\textsc{ILP}\xspace}

\newcommand{\ecbs}{\textsc{ECBS}\xspace}

\title{
Spatial and Temporal Splitting Heuristics for Multi-Robot Motion Planning 
\vspace*{-2mm}
}
\author{
Teng Guo \quad  Shuai D. Han \quad Jingjin Yu
\thanks{
G. Teng, S. D. Han, and J. Yu are with the Department of Computer Science, Rutgers, 
the State University of New Jersey, Piscataway, NJ, USA. 
E-Mails: \{{\tt teng.guo, shuai.han, jingjin.yu}\}\hspace*{.25em} \MVAt \hspace*{.25em}{\tt rutgers.edu}. 
This work is supported in part by NSF award IIS-1734419 and IIS-1845888.
}
}

\begin{document}

\maketitle

\thispagestyle{empty}
\pagestyle{empty}

\begin{abstract}
In this work, we systematically examine the application of spatio-temporal splitting heuristics to the Multi-Robot Motion Planning (MRMP) problem in a graph-theoretic setting: a problem known to be NP-hard to optimally solve. Following the divide-and-conquer principle, we design multiple spatial and temporal splitting schemes that can be applied to any existing MRMP algorithm, including integer programming solvers and Enhanced Conflict Based Search, in an orthogonal manner. The combination of a good baseline MRMP algorithm with a proper splitting heuristic proves highly effective, allowing the resolution of problems $10+$ times than what is possible previously, as corroborated by extensive numerical evaluations. Notably, spatial partition of problem fusing with the temporal splitting heuristic and the enhanced conflict based search (ECBS) algorithm increases the scalability of ECBS on large and challenging DAO maps by $5$--$15$ folds with negligible impact on solution optimality. 
\end{abstract}

\section{Introduction}\label{sec:intro}
We study the labeled Multi-Robot Motion Planning problem (\mpp) under a 
graph-theoretic setting, also known as Multi-Agent Path Finding (MAPF). 
The basic objective of \mpp is to find a set of collision-free paths to route multiple robots from a start configuration to a goal configuration.  
In practice, solution optimality is also of key importance; yet optimally solving \mpp is generally NP-hard~\cite{YuLav13AAAI,Sur10,Yu2015IntractabilityPlanar}. 
As one can readily imagine, given the ubiquity of the problem setting, effective algorithms find many important large-scale applications, e.g., warehouse automation~\cite{WurDanMou08,guizzo2008three}. Other application scenarios include formation~\cite{PodSuk04, SmiEgeHow08}, agriculture~\cite{cheein2013agricultural}, object 
transportation~\cite{RusDonJen95}, swarm robotics~\cite{preiss2017crazyswarm}, to list a few. 
Due to the wide range of impactful applications, even though \mpp had been studied since the 1980s in the robotics domain~\cite{KorMilSpi84,ErdLoz86,LavHut98b,GuoPar02}, it remains an active research topic. 
Many effective algorithms, for example~\cite{YuLav16TRO, boyarski2015icbs, cohen2016improved}, have been proposed recently that balance fairly well between computational efficiency and solution optimality. 

Nevertheless, there persists the practical need to continuously improve the efficiency and scalability of \mpp solutions, since a few percentage of computation time or path quality difference on path planning and motion execution could significantly affect the efficiency and throughput of these multi-robot systems. Such needs motivate us to carefully examine three key factors in \mpp that impact the performance of related algorithms: \emph{the number of robots} $n$, \emph{the size and complexity of the environment} $S$,
%
and \emph{the planning horizon} $T$. The overall complexity of a given problem can be measured as a function of these three factors, i.e., $f(n, S, T)$. A simple but reasonable approximation is the product $f(n, S, T) \propto nST$. The measure is rough as the factors are interrelated. For example, as $n/S$ approaches its upper limit for a given $S$ (i.e., the robot density hitting extremes), the complexity can grow exponentially in $n$. 
Most existing methods for optimally solving \mpp work with one or more of these factors. The most popular \emph{decoupling} approach~\cite{ErdLoz86,standley2010finding,wagner2011m,sharon2015conflict} essentially treats each robot individually, handling interactions on an ad-hoc basis. Spatial and temporal domains have also been exploited, though rather sparsely. In~\cite{YuLav16TRO}, a rudimentary divide-and-conquer approach is applied to split the planning horizon to $2^m$ slices, decoupling a problem over the time domain. Combined spatial-temporal approach has also been exploited, e.g., in~\cite{silver2005cooperative}, where a space-time window is used to reduce the computational effort. 

In this work, we made a first attempt to systematically exploit the application of divide-and-conquer over spatial and temporal domains, at the global level. Careful spatial and/or temporal division can be applied to most existing \mpp algorithms in an orthogonal manner, often bringing significant performance boosts. Specifically, our \textbf{main contributions} are: 
{\em (i)} We exploit multiple schemes for decoupling an \mpp instance over the temporal domain. Combined with effective solvers such as ILP~\cite{YuLav16TRO} or ECBS~\cite{barer2014suboptimal}, problems with many folds more robots can be readily solved, often with minimal impact on the solution optimality.
{\em (ii)} We devise schemes for decoupling an \mpp instance over the spatial domain that is also compatible with temporal decoupling schemes. Spatial division heuristics allow much larger \mpp instances to be solved without significant impacts on the solution optimality.

\textbf{Related Work.} 
Whereas the feasibility question has been answered for \mpp~\cite{KorMilSpi84}, due to the hardness~\cite{Sur10,YuLav13AAAI,Yu2015IntractabilityPlanar}, many attempts have been made at optimally solving \mpp 
\cite{hart1968formal,ErdLoz86,Sil05,standley2010finding,wagner2011m,sharon2015conflict,sharon2013increasing,yu2018constant}. 
Among these, combinatorial-search based solvers have been demonstrated to be effective. 
One of the earliest work is Local Repair A${}^*$ (LRA${}^*$)~\cite{hart1968formal}, which employs a basic form of decoupled search assisted with local repairs. The decoupling idea was also explored in~\cite{ErdLoz86}. Subsequently, a windowed approach~\cite{Sil05} was shown to provide additional efficiency gain through restricting the spatio-temporal search domain. Specific heuristics were later developed, including independence detection~\cite{standley2010finding}, sub-dimensional expansion~\cite{wagner2011m}, conflict-based search~\cite{sharon2015conflict}, increasing-cost-tree search~\cite{sharon2013increasing}, to list a few. 
It is also possible to solve the problem through reducing \mpp to other problems, e.g., SAT~\cite{surynek2012towards}, answer set programming~\cite{erdem2013general}, integer linear programming (ILP)~\cite{YuLav16TRO}.
Though these converted problems are also hard, they in fact facilitate the optimal resolution of the original \mpp problem due to the availability of specialized solver. 
As optimal solvers can be time consuming to run, sub-optimal solutions to \mpp have 
also be extensively studied. Solvers like push and swap~\cite{luna2011push}, push and rotate~\cite{de2013push}, windowed hierarchical cooperative A${}^*$, developed as part of~\cite{silver2005cooperative}, all return feasible solutions quickly. Balancing efficiency and optimality
is one of the most attractive topics; some algorithms emphasize scalability without sacrificing much optimality, e.g., enhanced conflict based
search (ECBS)~\cite{barer2014suboptimal}, DDM~\cite{han2020ddm}. 

Relating to our work, divide and conquer techniques had been applied
to tackle optimally solving \mpp. Similar to approaches explored in this work, 
sub-goals and sub-plans are stitched together to construct a global plan for 
multi-agent planning in~\cite{ephrati1994divide}, which reduces the branching 
factor, leading to reductions in computation time. The {\em{$k$-way-split}} \ilp~\cite{YuLav16TRO} divides a problem into equal sized sub-problems 
by finding intermediate goals in the middle of individual paths. 
It effectively reduces the computation time of the \ilp solver. The time domain split heuristic in this work builds on these earlier ideas and renders 
them more general. 


\textbf{Organization.} 
In Section~\ref{sec:preliminaries}, we formally define the multi-robot motion planning problem and provide preliminaries for ILP and ECBS. 
In Section~\ref{sec:time} and Section~\ref{sec:space} we describe the time split heuristic and space heuristic respectively.
In  Section~\ref{sec:experiment},  we  provide  evaluation  results  of these heuristics combined with \mpp solvers.  We conclude in Section~\ref{sec:conclusion}.

\section{Preliminaries}\label{sec:preliminaries}

The {\em Multi-Robot Motion Planning} problem (\mpp) is defined on an undirected graph $G = (V, E)$. 
We assume that $G$ is a grid graph by default. 
That is, given integers $w$ and $h$ as the graph's {\em width} and {\em height}, 
the vertex set can be represented as $V \subseteq \{(i, j) \mid 1 \leq i \leq w, 1 \leq j \leq h,i\in \mathbb{Z},j\in \mathbb{Z}\}$. 
The graph is $4$-way connected, i.e., for a vertex $v = (i, j)$, the set of its neighboring 
vertices are defined as $N(v) = \{(i + 1, j),(i - 1, j),(i, j + 1),(i, j - 1)\} \bigcap V$. 
The problem involves $n$ robots $r_1, \dots, r_n$, 
where each robot $r_i$ has a unique start state $s_i \in V$ and a unique goal state $g_i \in V$. 
We denote the joint start configuration as $X_S = \{s_1, \dots, s_n\}$ and 
the goal configuration as $X_G = \{g_1, \dots, g_n\}$. 
The objective of \mpp is to find a set of feasible path for all robots. 
Here, a {\em path} for robot $r_i$ is defined as a sequence of $T + 1$ vertices 
$P_i = (p_i^0, \dots, p_i^T)$ that satisfies: 
{\em (i)} $p_i^0 = s_i$; 
{\em (ii)} $p_i^T = g_i$; 
{\em (iii)} $\forall 1 \leq t \leq T, p_i^{t - 1} \in N(p_i^t)$. 
Apart from the feasibility of each individual path, for $P$ to be collision-free, 
$\forall 1 \leq t \leq T, 1 \leq i < j \leq n$, $P_i, P_j$ must satisfy 
{\em (i)} $p_i^t \neq p_j^t$ (no collisions on vertices); 
{\em (ii)} $(p_i^{t - 1}, p_i^t) \neq (p_j^t, p_j^{t - 1})$ (no head-to-head collisions on edges). 

In this work, we consider two optimization objectives. 
The first objective is to minimize the {\em makespan}, which is the time for all robots to reach the 
goal vertices. Following our problem definition, the makespan objective is interpreted as $\min T$. 
The second objective is to minimize the {\em sum-of-costs}, a cumulative cost function that sums over 
all robots of the number of time steps required to reach the goals. For each robot, denoting $t_i$ 
such that $\forall t_i \leq t \leq T, p_i^t = g_i$, the sum-of-costs objective is calculated as 
$\min \sum_{1 \leq i \leq n} t_i$. We point out that this later objective is also often a good proxy to the total travel distance objective. 

The problems studied in this work are as follows.

\vspace*{-2mm}
\begin{problem}
{\normalfont \bf Min-Makespan \mpp.}
Given $(G, X_{S}, X_{G})$, find a conflict-free path set $P$ that routes the robots from $X_{S}$ to $X_{G}$ and minimizes makespan $T$.  
\end{problem}
\vspace*{-4mm}
\begin{problem}
{\normalfont \bf Min-Sum-of-Costs \mpp.}
Given $(G, X_{S}, \allowbreak X_{G})$, find a conflict-free path set $P$ that routes the robots from $X_{S}$ to $X_{G}$ and minimizes sum of costs $\sum_{1 \leq i \leq n} t_i$.  
\end{problem}
\vspace*{-2mm}




Instead of developing full algorithms, this work focuses on 
heuristics for dividing an \mpp instance into sub-problems. These are solved using existing algorithms, in parallel when possible. The 
two classes of heuristics divide the original problem in time or space domain. 
After the split, we make sure that the solution for one sub-problem does not 
affect computing solutions for the others, thus maintaining the completeness 
guarantee of the existing \mpp algorithms.


\section{Splitting Over the Time Domain}\label{sec:time}

Our time-division heuristic has its roots in a split heuristic from~\cite{YuLav16TRO}. We first provide a brief introduction of that heuristic, and continue to describe our significant generalizations. 

\begin{figure}[ht!]
    \centering
    \subfigure[]{
        \centering
        \begin{overpic}[width=0.47\linewidth]{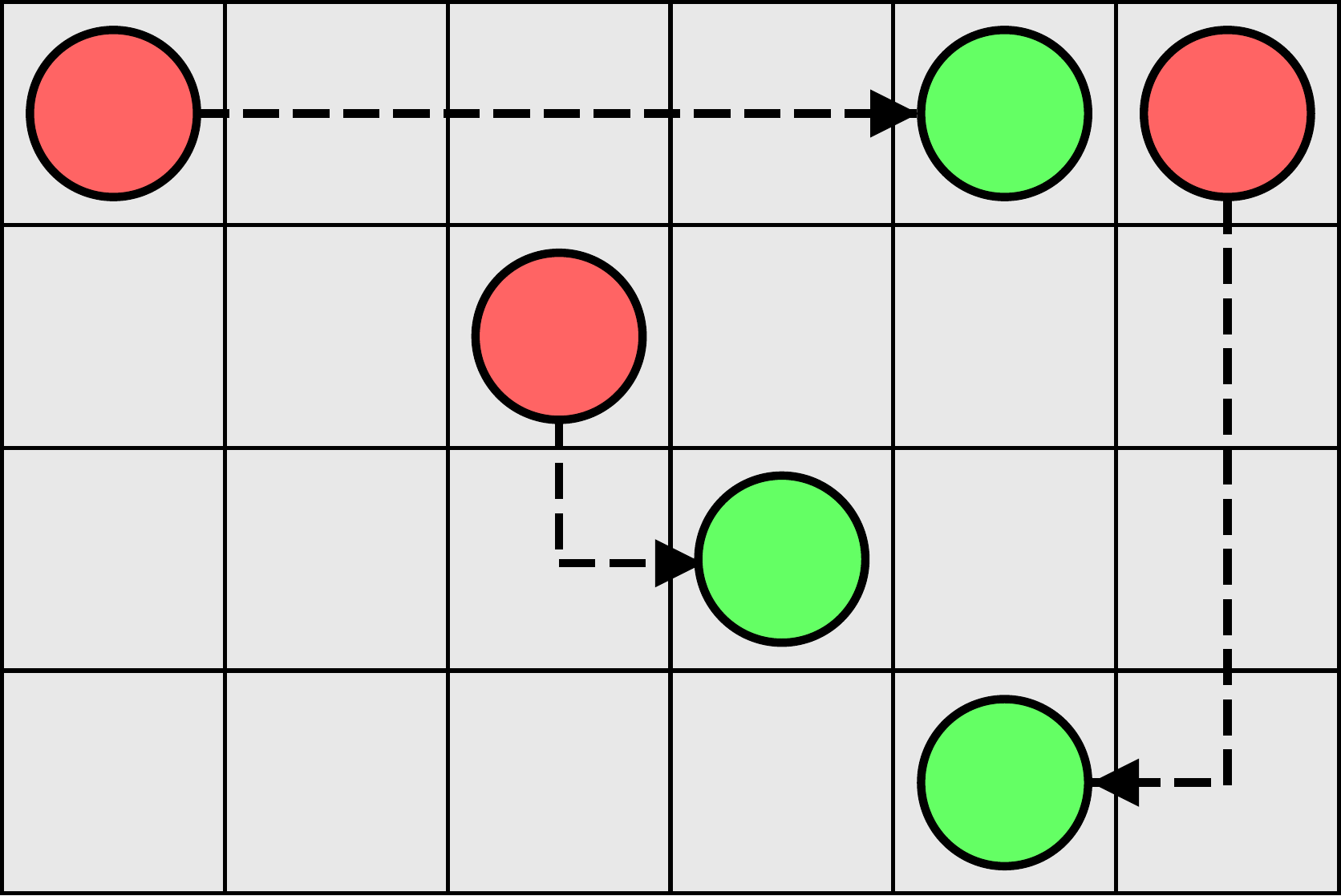}
          \put (5, 57) {$s_1$}
          \put (38, 40.25) {$s_2$}
          \put (88, 57) {$s_3$}
          \put (71.5, 57) {$g_1$}
          \put (54.5, 24) {$g_2$}
          \put (71.5, 7.5) {$g_3$}
        \end{overpic}
   }\hfill
       \subfigure[]{
        \centering
        \begin{overpic}[width=0.47\linewidth]{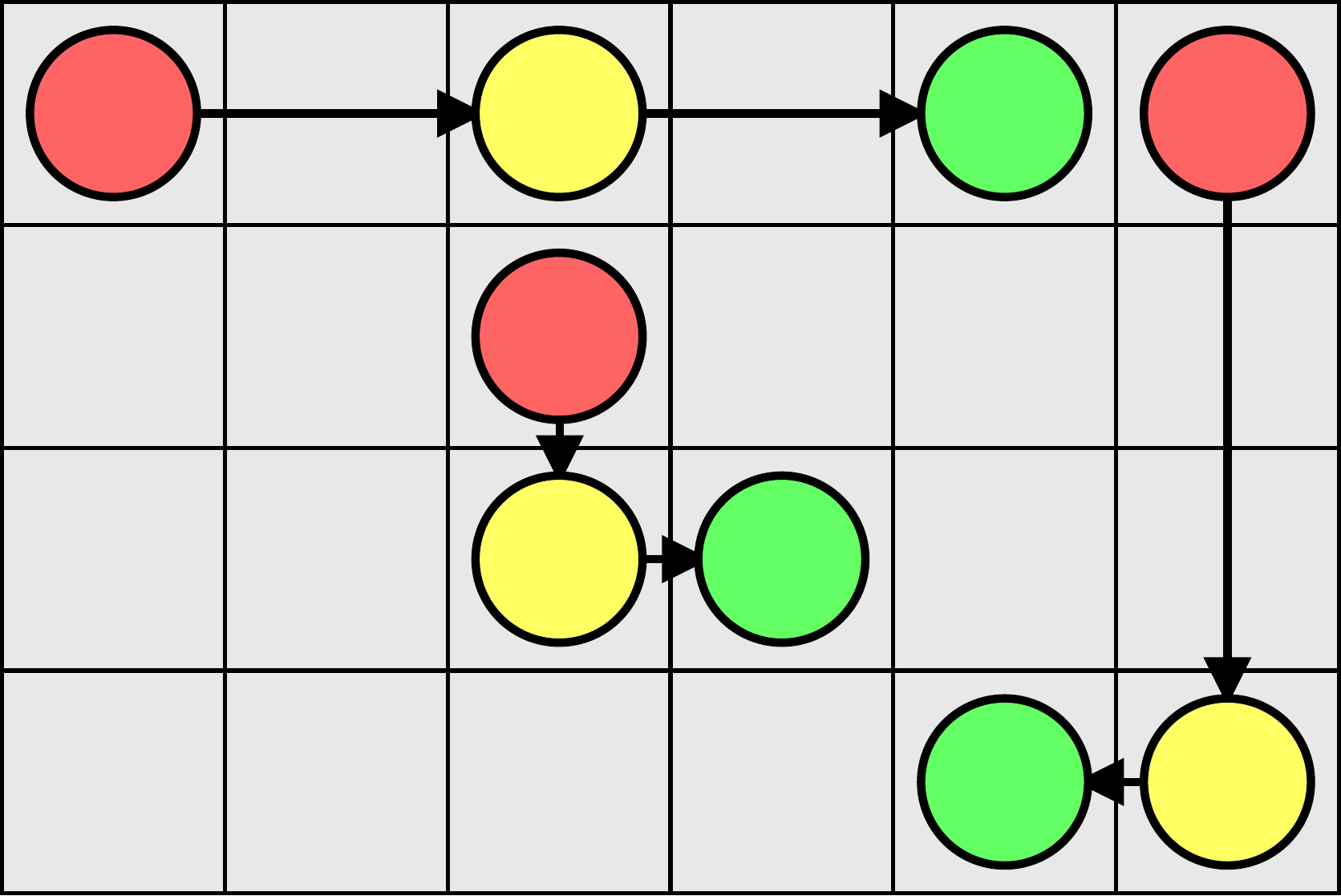}
          \put (5, 57) {$s_1$}
          \put (38, 40.25) {$s_2$}
          \put (88, 57) {$s_3$}
          \put (71.5, 57) {$g_1$}
          \put (54.5, 24) {$g_2$}
          \put (71.5, 7.5) {$g_3$}
        \end{overpic}
      }
    \caption{\label{fig:time-split-example} 
    An example of the time-split heuristic. 
    (a) The original problem in a $6\times4$ grid with $3$ robots. 
    The start and goal configurations are visualized using red and green disks. 
    The dashed lines show one set of possible solution paths. 
    (b) With the time-split heuristic, the problem can be solved in two phases. 
    The robots are first moved to an intermediate configuration (yellow disks), 
    and then to the goals. 
    }
\end{figure}
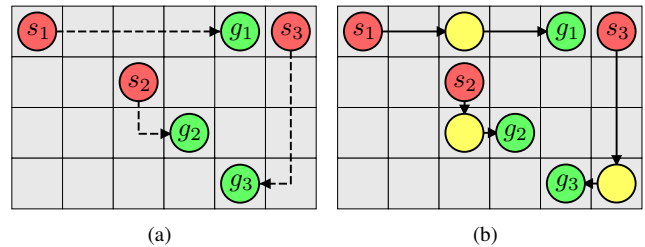

Given an \mpp instance, the original $k$-way ($k$ as a power of $2$) split heuristic~\cite{YuLav16TRO} divides a problem into $k$ equal sized 
sub-problems. Denoting the original start configuration as $X_S$ and goal configuration as $X_G$, in the first iteration, the heuristic finds an \emph{intermediate configuration} $X_{IM}$, and generates one sub-problem that routes robots from $X_S$ to $X_{IM}$, and another sub-problem that routes robots from $X_{IM}$ to $X_G$. Here, for each robot, its intermediate configuration in $X_{IM}$ is a vertex that is roughly the same distance to the robot's start and goal vertices. Such a process is then recursively applied to the two sub-problems for another $k/2-1$ times each. After all sub-problems are created and solved individually, a solution to the original problem is found by concatenating solutions for the sub-problems. 
An example of such a time division is provided in Fig.~\ref{fig:time-split-example}. 

\subsection{Arbitrary Splitting over the Time Domain}

The original $k$-way split limits $k$ to be powers of $2$ and force each sub-problem to have roughly equal underestimated makespan (i.e. the minimum possible makespan when ignoring robot collisions). 
We remove these limitations, allowing splitting the original problem into arbitrary number of sub-problems with different underestimated makespan. 

Given an \mpp instance $(G, X_S, X_G)$ and $k\in \mathbb{Z}$, our time-split 
heuristic (Algorithm~\ref{alg:k-time}) first computes a shortest path $P_i$ for every robot $i$. Each time the instance is split, conflict-free intermediate states are located. 
For each intermediate state $x_{IM}^{ij}$, the intermediate state for robot $i$ in the $j$-th sub-problem, we consider all robots in a descending 
order of the shortest path length.
Then for each robot $i$, we find a series of candidate intermediate goal states $X_{IM}^{i}$. 

The intermediate state $x_{IM}^{ij}$ is picked from the vertices that are about 
$d_{ij}=j\cdot \text{dist}(x_{S}^{i},x_{G}^{i}) /k$  from the start vertex and $|P_i|-d_{ij}$ from the goal vertex while avoiding conflicts, where $\text{dist}(u,v)$ is denoted as the shortest distance between two vertices $u$ and $v$.
If no conflict-free intermediate state is found, vertices whose distance form start vertex are $d_{ij}\pm 1$ are considered. 
When all intermediate goal states are decided, a polynomial time algorithm (i.e.~\cite{kornhauser1984coordinating,khorshid2011polynomial}) $CheckSolvable()$ 
can check if the resulting sub-problems are solvable.
This procedure is repeated until a feasible intermediate goal state is found.
The final feasible intermediate states denoted as $X_{IM}:=\{x_{IM}^{ij}; 1 \leq j \leq k - 1,1\leq i\leq n\}$  will be returned.
In this way, the initial instance is split into $k$ sub-problems, $P_{1}(G,X_{S},X_{IM}^{1}),...,P_{k}(G,X_{IM}^{k-1},X_{G})$. 
Any \mpp solvers may be applied to solve the resulting sub-problems. 
Since there is no interaction between the individual sub-problems, once we obtain the solution for each sub-instance the final path can be obtained by concatenating them together. 
The final makespan is obtained by adding all the makespan of each sub-problem together, which is $T=\sum_{j}T_{j}$. 
In practice, the simple heuristic dramatically improves  algorithm  performance  without  heavy  negative  impact  on  path  optimality  in  terms  of  makespan;  we  observe  a consistent speedup in computational experiments.

\begin{algorithm}
\begin{small}
  \caption{{\em{$k$-time-split}\label{alg:k-time}} for {\em{min-makespan}} \mpp}
  \KwIn{Start and goal configurations $X_{S}$, $X_{G}$, graph $G$}
  Call A* to find an individual path $P_{i}$ for each robot $i$\;
  Sort paths $P_i$ according to path length in descent order\;
  
  \For{$j=1$ to $k-1$}
  {
  \While{true}{
  $H_{used}\leftarrow \emptyset$\;
  \For{$i=1$ to $n$}
  {
    $S_{1}\leftarrow \emptyset$, $S_{2}\leftarrow\emptyset$, $d_{ij}\leftarrow j\cdot dist(x_{S}^{i},x_{G}^{i}) /k$\;
    $s_{min}=s_{max}=d_{ij}$, $g_{min}=g_{max}=|P_{i}|-d_{ij}$\;
    \While{true}
    {
      Find the vertices whose distance is in $[s_{min},s_{max}]$ from $x_{S}^{i}$, add them to $S_{1}$\;
      Find the vertices whose distance is in $[g_{min},g_{max}]$ from $x_{G}^{i}$,  add them to $S_{2}$\;
      $V\leftarrow(S_{1}\cap S_{2})-H_{used}$\;
      \If{$V\neq \emptyset$}
      {
         Choose random $x_{IM}^{ij}\in V$, add it to $H_{used}$\;
         break\;
      }
      \Else{
        $s_{min}\leftarrow s_{min}-1,s_{max}\leftarrow s_{max}+1$\;
        $g_{min}\leftarrow g_{min}-1,g_{max}\leftarrow g_{max}+1$\;
        }
    }
 }
    \lIf{$CheckSolvable(X_{IM}^{j})$}{break} 
    }
  }
  return $X_{IM}$\;
\end{small}
\end{algorithm}

As a further generalization, our time-split also allows splitting the problem into 
instances with arbitrary ratio.
For arbitrary ratio $\lambda_{j}$ with $\sum_{j}\lambda_{j}=1$, we just let $d_{ij}=(\sum_{\ell = 1}^{\ell=j}\lambda_{\ell}) dist(x_{S}^{i},x_{G}^{i})$.
For {\em{min-makespan}} \mpp, if we decide to split original problem into $k$ sub-problems, we observe that even splits are generally better than uneven splits. 
Empirically, the computational time of \mpp solvers is largely determined by the time span. 
Since the computational time of time-split \mpp is decided by the maximum running time to 
solve each sub-problem, even splits lead to the smallest expected maximum time span of sub-problems, 
and consequently make the parallelization more efficient.


\subsection{Special Considerations for Min-Sum-of-Cost Objective}

Apart from requiring $k$ to be a power of $2$, the original $k$-way split 
heuristic performs poorly in solving {\em min-sum-of-costs} \mpp. 
Since the heuristic generates intermediate configurations by equally splitting the shortest paths, robots often cannot reach the goal configurations until the last sub-problem. Thus, a robot does not reach the goal vertex as fast as possible, even though it might be very close to the goal in the beginning. 
As an example, suppose that a 2-way split is carried out with each sub-problem having a time horizon of $T/2$. If a robot $r_{i}$ does not move in the solution to the second sub-problem (i.e., $T/2\leq t\leq T$), it contributes 0 to the total distance. 
However, if $r_{i}$ moves even a single step in the solution to the first sub-problem (i.e., $0 \leq t \leq T/2$), then $r_{i}$ will contribute at least $T/2$ to the total sum of costs. 
Thus, the final sum of costs obtained would be highly sub-optimal as pure an artifact of the heuristic.


We modify the {\em{min-makespan}} version of time-split, making it applicable to {\em{min-sum-of-costs}} \mpp.
Take 2-way split as an example, suppose that the makespan lower bound of original \mpp is $T$, we still break it into two sub-problems with time horizon of $T/2$ each. 
Instead of choosing intermediate states at the middle of each individual path, for robot $r_{i}$ the vertex whose distance is $d_{i}=\min(T/2,|P_{i}|)$ from start vertex while $|P_{i}|-d_{i}$ from goal vertex would be chosen, where $|P_{i}|$ is the path length of robot $r_{i}$ found by A* ignoring conflicts with other robots. By setting $T/2$ as the threshold time-span, robots can reach their goal as fast as possible and the resulting sum of costs lower bound of the two sub-problems are additive. 



\vspace*{-2mm}
\begin{lemma}
If the original problem is feasible, the time-split heuristic always generates feasible sub-problems.
\end{lemma}
\vspace*{-2mm}
\begin{proof}
Assume original problem $P(G,X_{S},X_{G})$ is solvable, then there must be an optimal solution 
$\mathbf{\Pi} = (\Pi_0, \dots, \Pi_T)$, with corresponding makespan $T$. 
Consider a configuration $\Pi_{j}=(\Pi_{1j},\Pi_{2j} \dots \Pi_{nj})$ where $\Pi_{ij}$ is denoted as the path vertex of robot $i$ at time step $j$, the sub-problems $P(G,X_{S},\Pi_{j})$ and $P(G,\Pi_{j},X_{G})$ are solvable. 
That is, a feasible problem indicates that feasible intermediate configurations always exist. 
Algorithm~\ref{alg:k-time} iterates over all of the possible configurations in the second outer loop and terminates in finite steps when a feasible configuration is found.
\end{proof}

\vspace*{-2mm}
\begin{proposition}
The time-split heuristic maintains the completeness of the existing \mpp algorithms.
\end{proposition}
\vspace*{-2mm}



\begin{remark}
\normalfont
Theoretically, time-split is complete on any graph. In the worst case, 
finding a feasible intermediate configuration takes $O(|V|^{n})$ time. However, in our 
experimental evaluation, when robot density is not extremely high, nearly every 
intermediate configuration found leads to solvable instances. Therefore, checking whether a sub-problem is solvable is unnecessary when robot density is not extremely high. 
\end{remark}

\vspace*{-2mm}
\begin{remark}
\normalfont
Time split heuristic is applicable in combination with any \mpp solvers; \ilp and \ecbs are chosen as representatives here. The performance of \ilp solver is heavily affected by the \ilp 
problem size, i.e. number of variables. Therefore, with smaller sub-problems to solve, 
time-split \ilp runs faster than non-split \ilp. As for \ecbs, in the worst case, 
the sub-problems adopt the whole map and expand all states, which is the same as 
non-split \ecbs. But in practice, the original problem is divided into sub-problems 
whose starts and goals are closer and it takes less time to find individual paths. 
Also, because the starts and goals are closer, usually when the instance is not very 
dense it takes less time to find solution for each sub-problem since there would be 
less conflicts in each sub-problem to resolve and thus the CT-tree needs to be 
searched is smaller. Besides, the heuristics allows us to take advantage of multiple 
cores and the resulting sub-problems can be solved in parallel. 
\end{remark}


\section{Splitting Over the Spatial Domain}\label{sec:space}

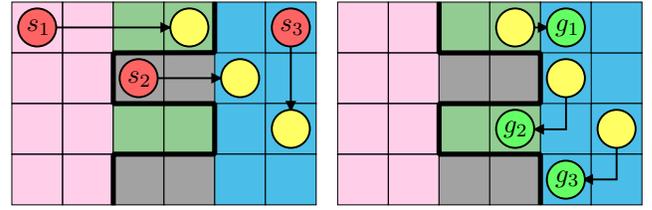
\begin{figure}[ht!]
    \centering
    \subfigure{
        \centering
        \begin{overpic}[width=0.47\linewidth]{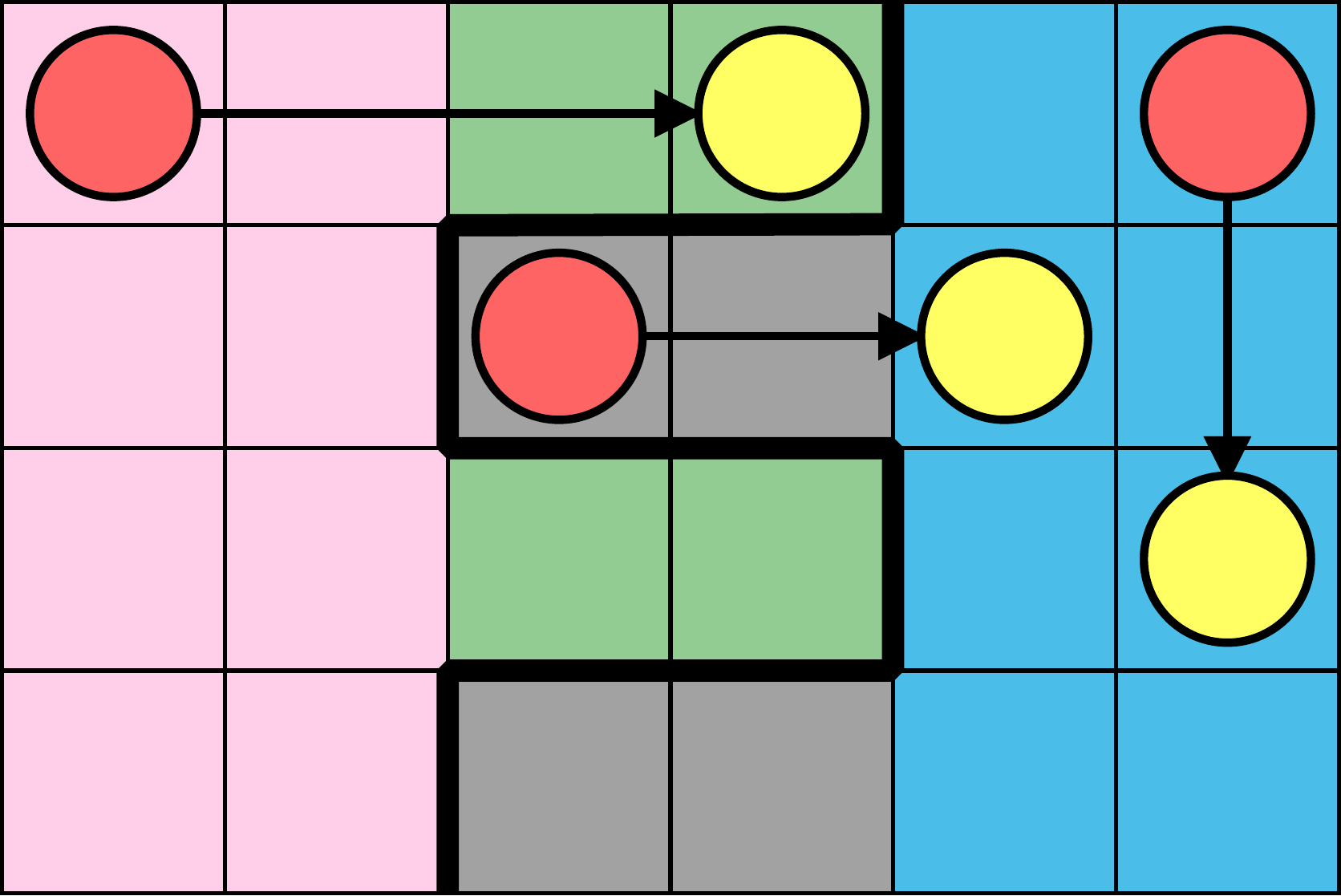}
          \put (5, 57) {$s_1$}
          \put (38, 40.25) {$s_2$}
          \put (88, 57) {$s_3$}
        \end{overpic}
        
   }\hfill
    \subfigure{
        \centering
        \begin{overpic}[width=0.47\linewidth]{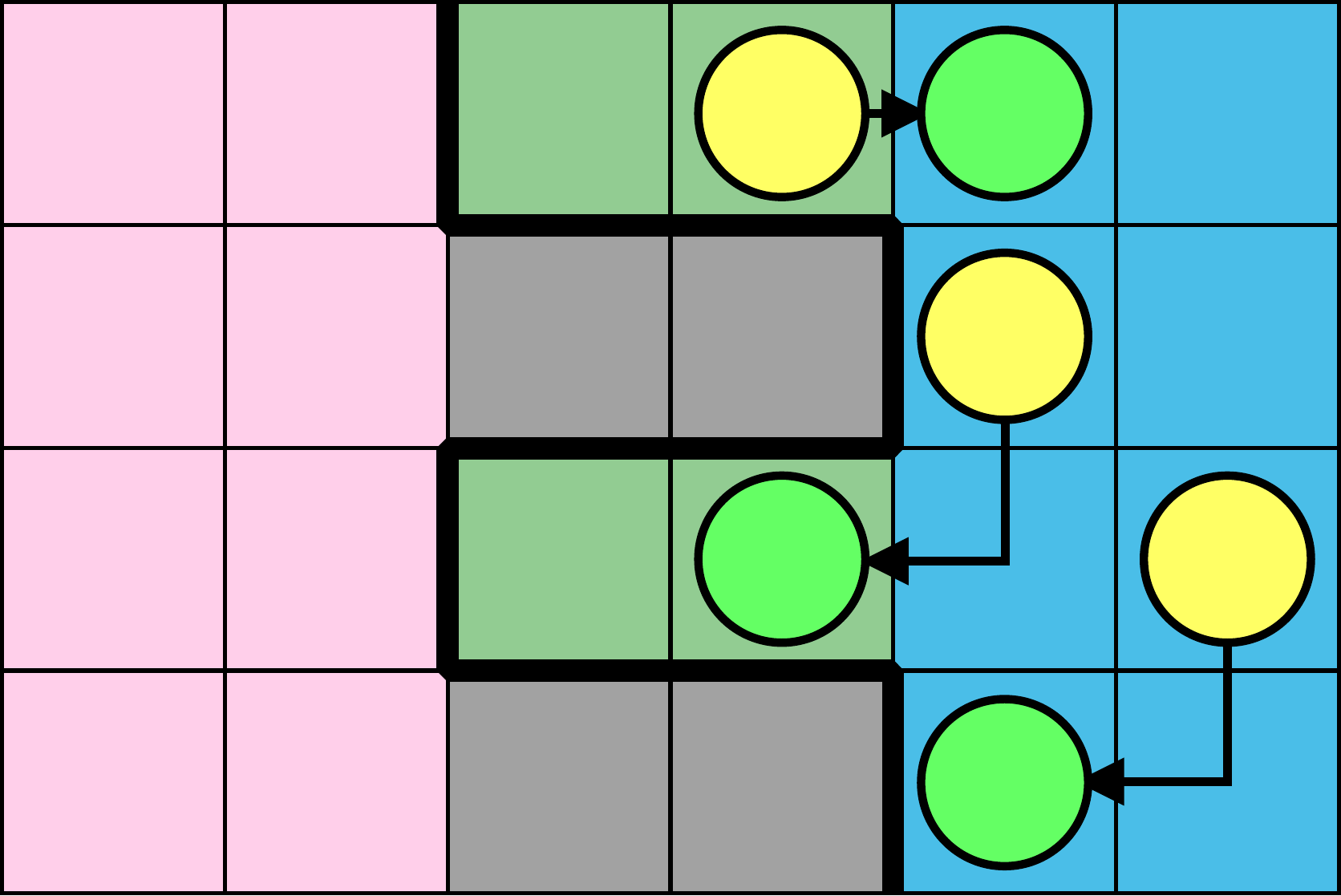}
          \put (71.5, 57) {$g_1$}
          \put (54.5, 24) {$g_2$}
          \put (71.5, 7.5) {$g_3$}
        \end{overpic}
     
    }
    \caption{\label{fig:sp-example} 
    Space split applied to the example in Fig.~\ref{fig:time-split-example}. 
    Here, $G_1, G_2, B_1, B_2$ are colored in pink, blue, green, grey, respectively. 
    The two sub-figures show the two sub-problems. 
    }
\end{figure}

As another natural route to the reduction of sub-problem sizes, a \emph{space-split} heuristic is explored which splits the original problem over the spatial domain. 
To stitch together the sub-problems, \emph{buffer zones} are introduced between divided regions of the environment (Fig.~\ref{fig:sp-example}). Essentially, buffer zones are regions with small blocks that allow robots to migrate from one region of the larger environment to another region of the environment between sub-problems. After examining multiple choices, we settled with buffer zones containing multiple small rectangular blocks that belong to different regions in different sub-problems.


As an illustration, for the instance in Fig.~\ref{fig:sp-example}, two buffer zones $B_{1}$ and $B_{2}$ are created. Each buffer zone contains two disconnected rectangular areas. The buffer zones separate the rest of the graph into two regions $G_{1},G_{2}$ where $G=G_{1}+B_{1}+G_{2}+B_{2}$. Here, we define operator ``$+$'' as $G(V,E)=G_{1}(V_1,E_1)+G_{2}(V_2,E_2)$ where $V=V_1\cap V_2$ and $E=\bigcap_{v\in V}\{(v,u)|u\in N(v)\}$.
Operator ``$-$'' is similarly defined. 


Depending on their starts and goals, robots are classified into 4 groups: (i) $x_{S} \in G_{1}+B_{1}$, $x_{G} \in G_{1}+B_{1}$; (ii). $x_{S}\in G_{2}+B_{2}$, $x_{G}\in G_{2}+B_{2}$; (iii). $x_{S}\in G_{1}+B_{1}$, $x_{G}\in G_{2}+B_{2}$;
 (iv). $x_{S} \in G_{2}+B_{2}$, $x_{G}\in G_{1}+B_{1}$.  
Depending on the classification, intermediate goal states are selected for each robot. For example, a robot going from $B_1$ to $B_2$ may require it to go into $G_1$ first. 
The rules of choosing intermediate states are described in Algorithms~\ref{alg:determine} and~\ref{alg:allocate}. Algorithm~\ref{alg:determine} shows how to classify the robots according to their starts and goals. If start and goal are in the same sub-graph, the intermediate state is chosen in that sub-graph.  If the goal is in another sub-graph and the robot is in the buffer zone, it should not be sent to the buffer zone. If the goal is in another sub-graph but not in the buffer zone, the robot should be sent to the buffer zone in the first phase.  Lines 7-8 ensure that the algorithm can always find a conflict-free intermediate goal even if the buffer zone is not large enough to hold all the robots.


\begin{algorithm}

\small
  \caption{ Determine intermediate state  \label{alg:determine}}
  \KwIn{Start $x_{S}$; goal $x_{G}$; subgraphs $G_{1},G_{2}$; buffer zones $B_{1},B_{2}$; set $H_{used}$}
  
  \If{$x_{S}$ and $x_{G}$ are in the same subgraph $G_{i}$}
  {
    $x_{IM} \leftarrow$  allocate($G_{i}$, $x_{S}$, $x_{G}$)\;
  }
  \ElseIf{$x_{G}$ is in another buffer zone}
  {
    $x_{IM} \leftarrow$ allocate($G_{i}-B_{i}$, $x_{S}$, $x_{G}$)\;
    \lElse{
    $x_{IM} \leftarrow$ allocate($B_{i}$, $x_{S}$, $x_{G}$)
    }
  }
  \lIf{$x_{IM}=null$}
  {
    $x_{IM}\leftarrow$ allocate($G_{i}$, $x_{S}$, $x_{G}$)
  }
  
  return $X_{IM}$\;

\end{algorithm}


\begin{algorithm}
\small
  \caption{ Allocate intermediate state \label{alg:allocate}}
  \KwIn{Start and goal $x_{S},x_{G}$;Subgraph $SG(V,E)$ ;Set $H_{used}$}
  $x_{IM}\leftarrow null$,  $minV\leftarrow +\infty$\;
  \For{$v\in V$}{
      $f\leftarrow \lambda_{1}(\max(dist(v,x_{S}),T_{1}) + \max(dist(v,x_{G}),T_{2})$\\
			$\qquad + \lambda_{2}\rho(v) + dist(v,x_{S}) + dist(v,x_{G})$\;
      \If{$f<minV$ and $ v \notin H_{used}$}
      {
        $x_{IM}\leftarrow v$\;
        $minV\leftarrow f$\;
      }
  }
  
  add $x_{IM}$ to $H_{used}$\;

  return $x_{IM}$\;

\end{algorithm}




Algorithm~\ref{alg:allocate} describes how intermediate states are chosen. 
We define $f$-value as a linear combination of the maximum makespan of two sub-problems, the local density $\rho(v)$, and the total distance a robot will travel. The local density $\rho(v)$ is defined as the number of occupied neighboring vertices of $v$.
The vertex in a given sub-graph with minimum $f$ value would be set as the intermediate goal.
For 2-split we use $T_{1}=T_{2}=T/2$ as the threshold to make sure that makespan of each sub-problem not exceed $T/2$ so that we can fully take the advantages of multi-core computation.

After all intermediate states are determined, the original problem is dealt with 
in two phases. In the first phase, robots are sent from starts to intermediate 
states and we need to solve $P_{11}(X_{S}^{(11)},X_{IM}^{(11)},G_{1}+B_{1})$,  $P_{12}(X_{S}^{(12)},X_{IM}^{(12)},G_{2}+B_{2})$. 
In the second phase, robots are sent from intermediate states to their goals 
and we need to solve $P_{21}(X_{IM}^{(21)},X_{G}^{(21)},G_{1}+B_{2})$, $P_{22}(X_{IM}^{(22)},X_{G}^{(22)},G_{2}+B_{1})$. 
Again, \ilp or any other general \mpp solvers can be readily applied to 
solve the resulting \mpp sub-problems in parallel.

In general cases, when applying {\em{space split}} to divide an original graph into $l\times m$ sub-graphs, we use a fixed buffer zone to complete the division. We find $k$ intermediate states for each robot and the solution procedure breaks into $k$ phases. In the $i$-th phase, the robots are sent from its $(i-1)$th intermediate state to $i$-th one. For each phase, there are $l\times m$ sub-instances need to solve and thus in total $l\times m\times k$ sub-instances need to be solved. The number of phases are determined by $l,m$ and the longest paths robots need to travel. Usually, $k$ is roughly $l+m$.
A major advantage of the space split is that the reduced environment size simultaneously induces a reduction in sub-problems' makespans, allowing the sub-problems to be easily solved. As such, scalability is significantly boosted.
The space-split can be combined with time-split, which brings further improvement to scalability. We denote the combination as {\em{time-space-split}}.

\section{Experimental Evaluation}\label{sec:experiment}

In this section, we evaluate how the proposed heuristics affect the {\em computation time} and the {\em optimality ratio}. The computation time is the time for an algorithm to generate a solution.The optimality ratio is measured as the solution cost over an underestimated cost, generated by moving all 
robots to the goals, ignoring collisions. For each test scenario, we push the number of robots up to the solvers' limit to test the effect of the heuristics on solvers' scalability. The start configuration, goal configuration and environment obstacles are uniformly randomly generated. Each result entry in this section is an average over $25$ test cases. 

We choose Integer Linear Programming~(\ilp)~\cite{YuLav16TRO} and Enhanced Conflict-Based Search~(\ecbs)~\cite{barer2014suboptimal} as the low-level \mpp solvers. These two algorithms are state-of-the-art in terms of solving \mpp on graphs. For \ecbs, we set its weight parameter $w = 1.5$ since it is a good balance between optimality and computational efficiency, as indicated in the original publication and from our observation. All experiments are executed on an Intel\textsuperscript{\textregistered} Core\textsuperscript{TM} i7-9700 CPU at 3.0GHz. Our heuristics are implemented in Java, while the \mpp solvers are in  C\nolinebreak[4]\hspace{-.05em}\raisebox{.4ex}{\small\bf ++}. 



\subsection{Evaluation of the Time Split Heuristic}

\begin{figure}[ht!]
  \centering
    \subfigure{
        \centering
        \includegraphics[width=0.465\linewidth]{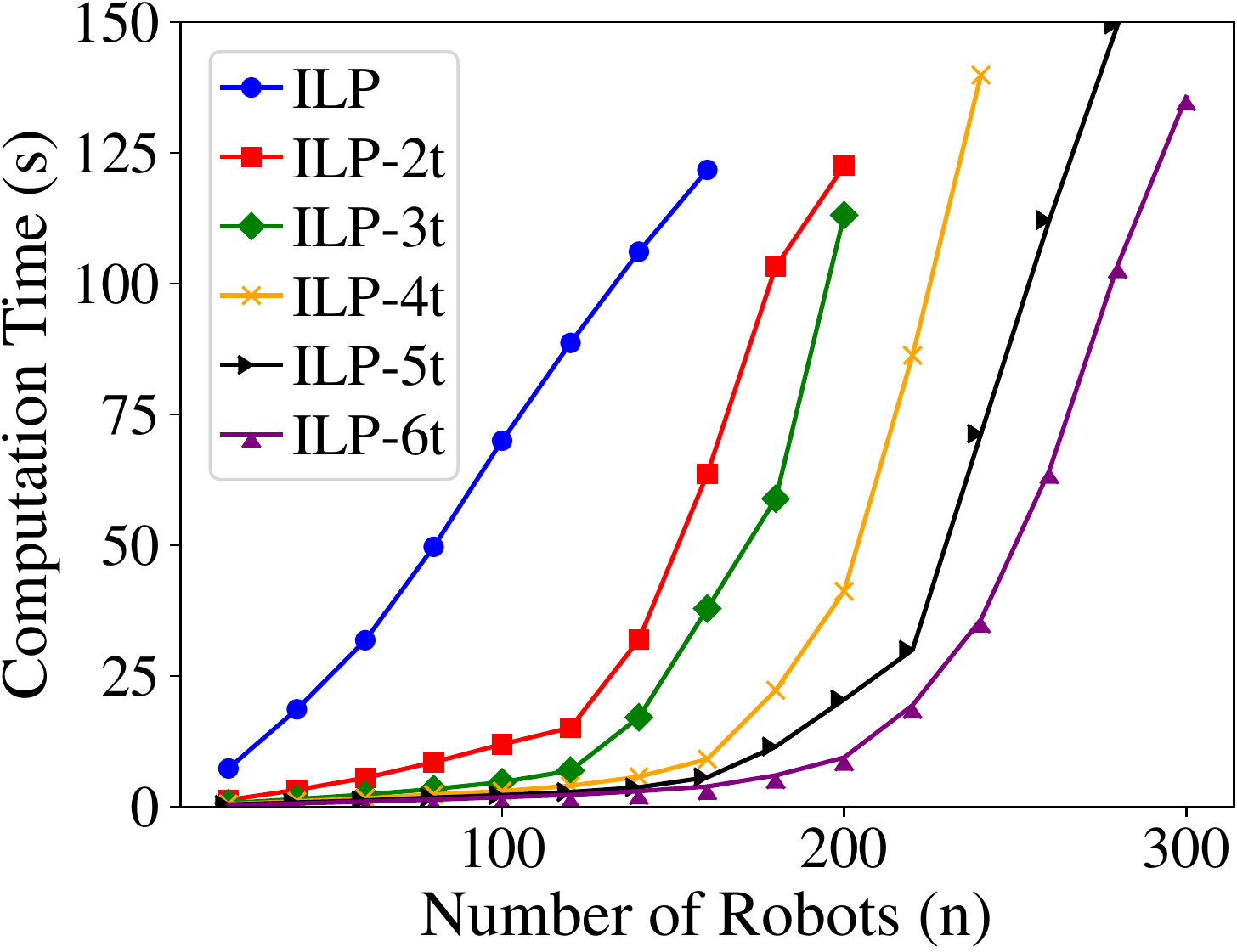}
    }
    \hfill
    \subfigure{
        \centering
        \includegraphics[width=0.465\linewidth]{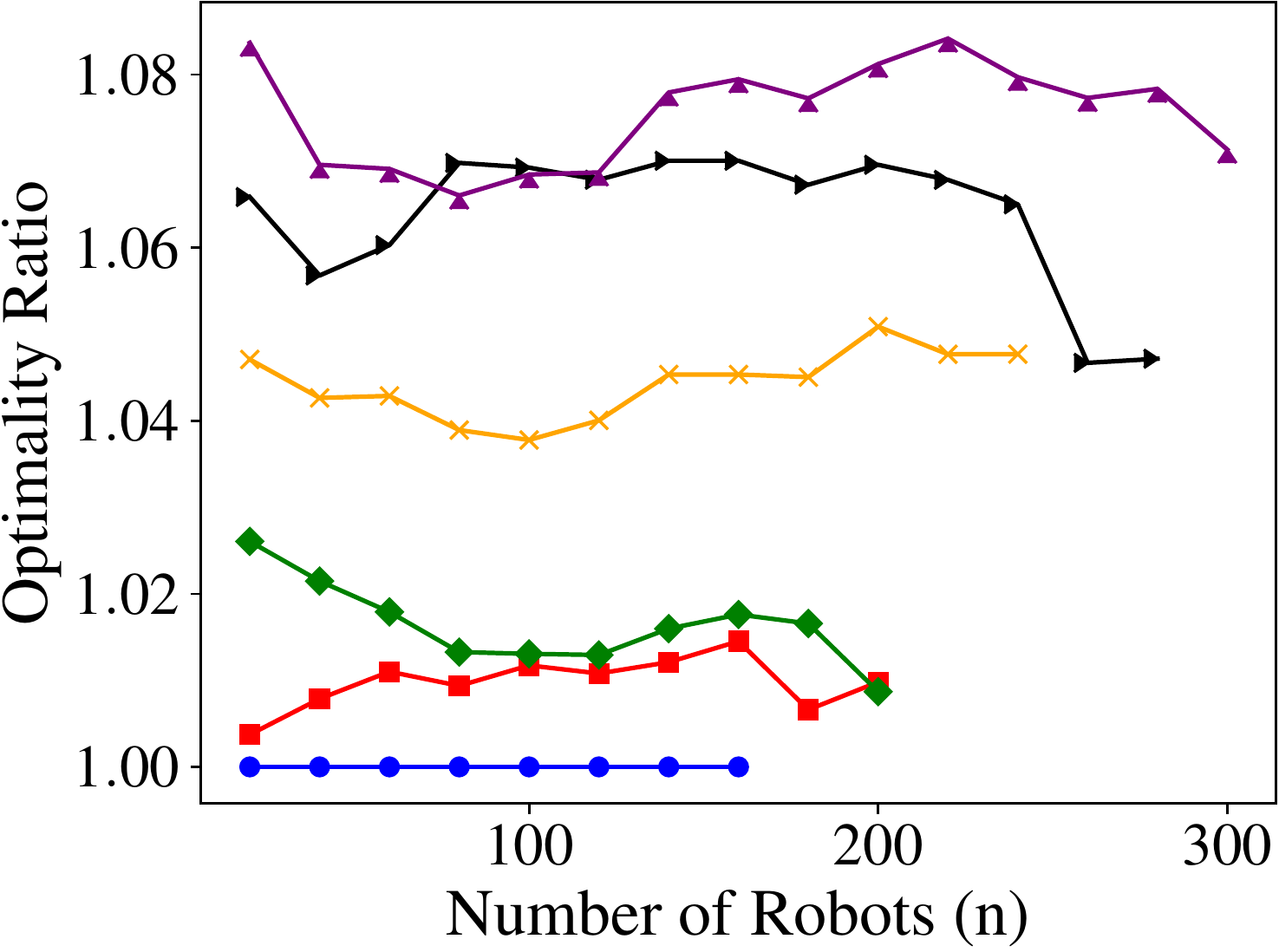}
    }
    \caption{\label{fig:time-ilp-makespan-32} 
    Result of {\em{$k$-time-split}} on {\em{min-makespan}} \mpp.}
\end{figure}

First, we evaluated $k$-time-split heuristic on a $32\times 32$ grid with $10\%$ obstacles. Fig.~\ref{fig:time-ilp-makespan-32} shows the makespan result using \ilp. Notation ILP-$kt$ stands for the combination of \ilp and $k$-time-split. We observe that the scalability of \ilp is significantly improved with minimal impact on solution optimality. For example, with 4-time-split, problems with $150$ robots can be solved around $25$ times faster, while the optimality ratio is well under $1.06$. 

\begin{figure}[ht!]
  \centering
    \subfigure{
        \centering
        \includegraphics[width=0.465\linewidth]{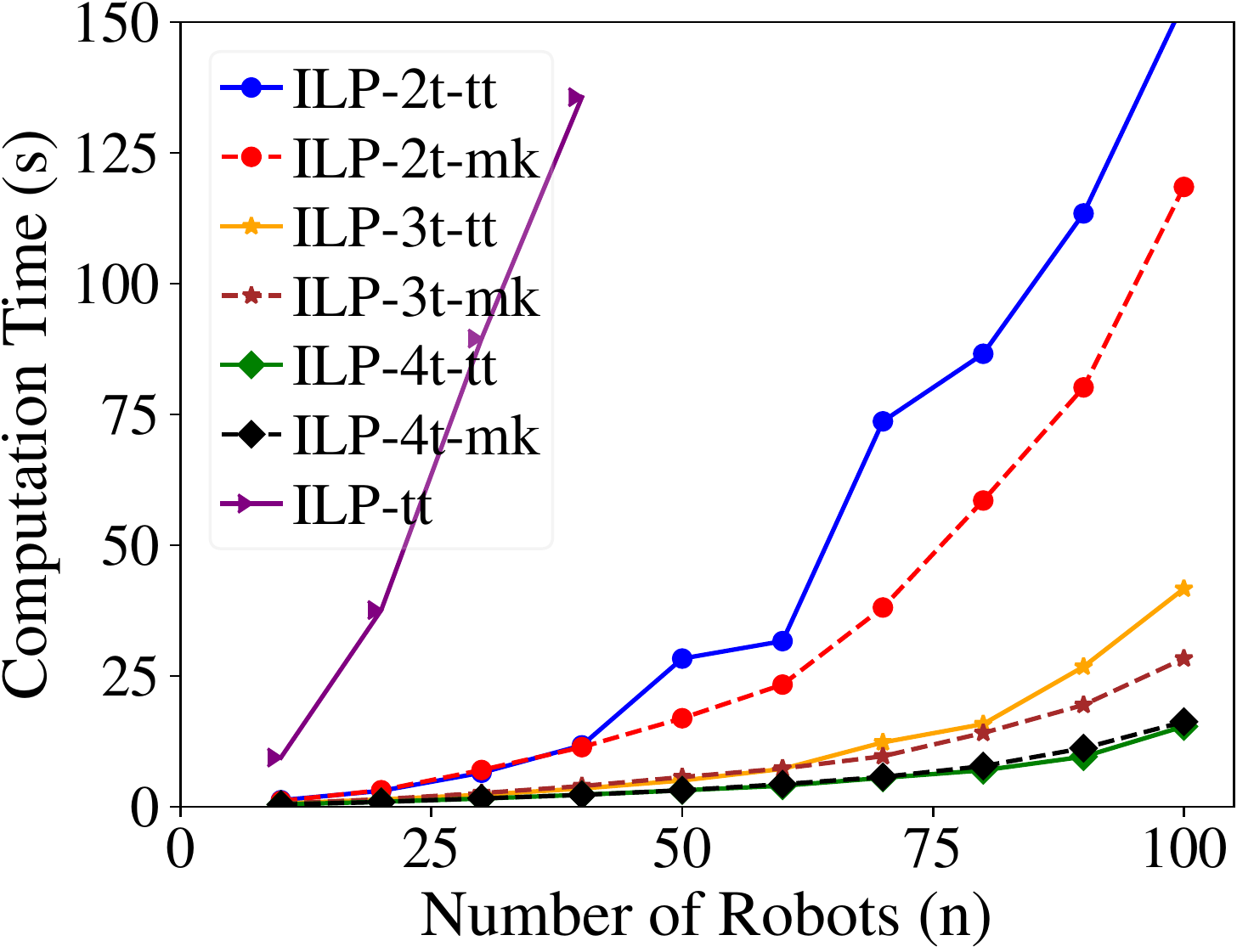}
    }
    \hfill
    \subfigure{
        \centering
        \includegraphics[width=0.465\linewidth]{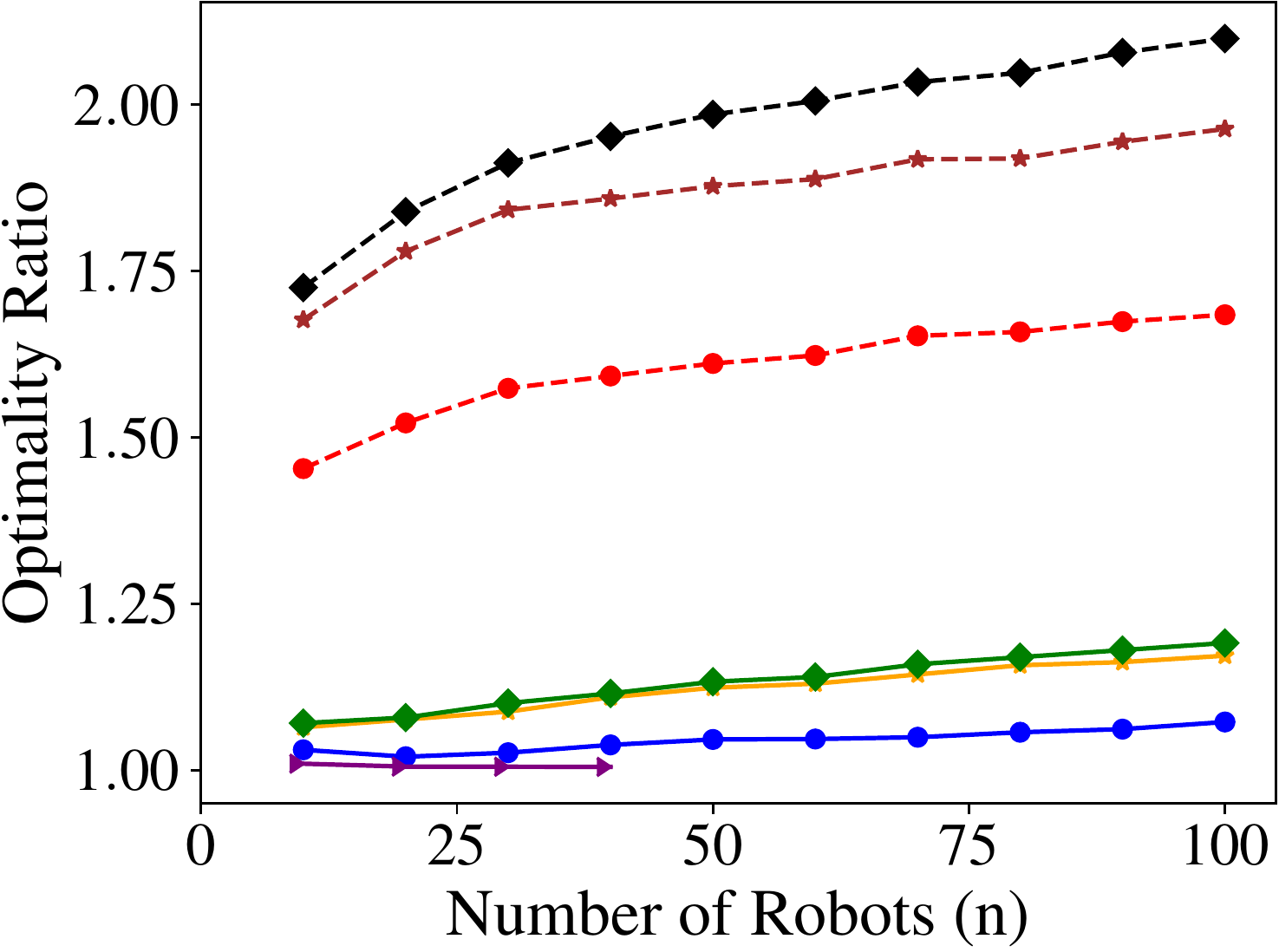}
    }
  \caption{\label{fig:time-ilp-sum-of-costs-32} 
    Comparison of makespan and sum-of-costs time-split heuristics on {\em min-sum-of-costs} \mpp. The test graph is $32\times 32$ grid with $10\%$ obstacles. }
\end{figure}

\begin{figure}[ht!]
  \centering
    \subfigure{
        \centering
        \includegraphics[width=0.465\linewidth]{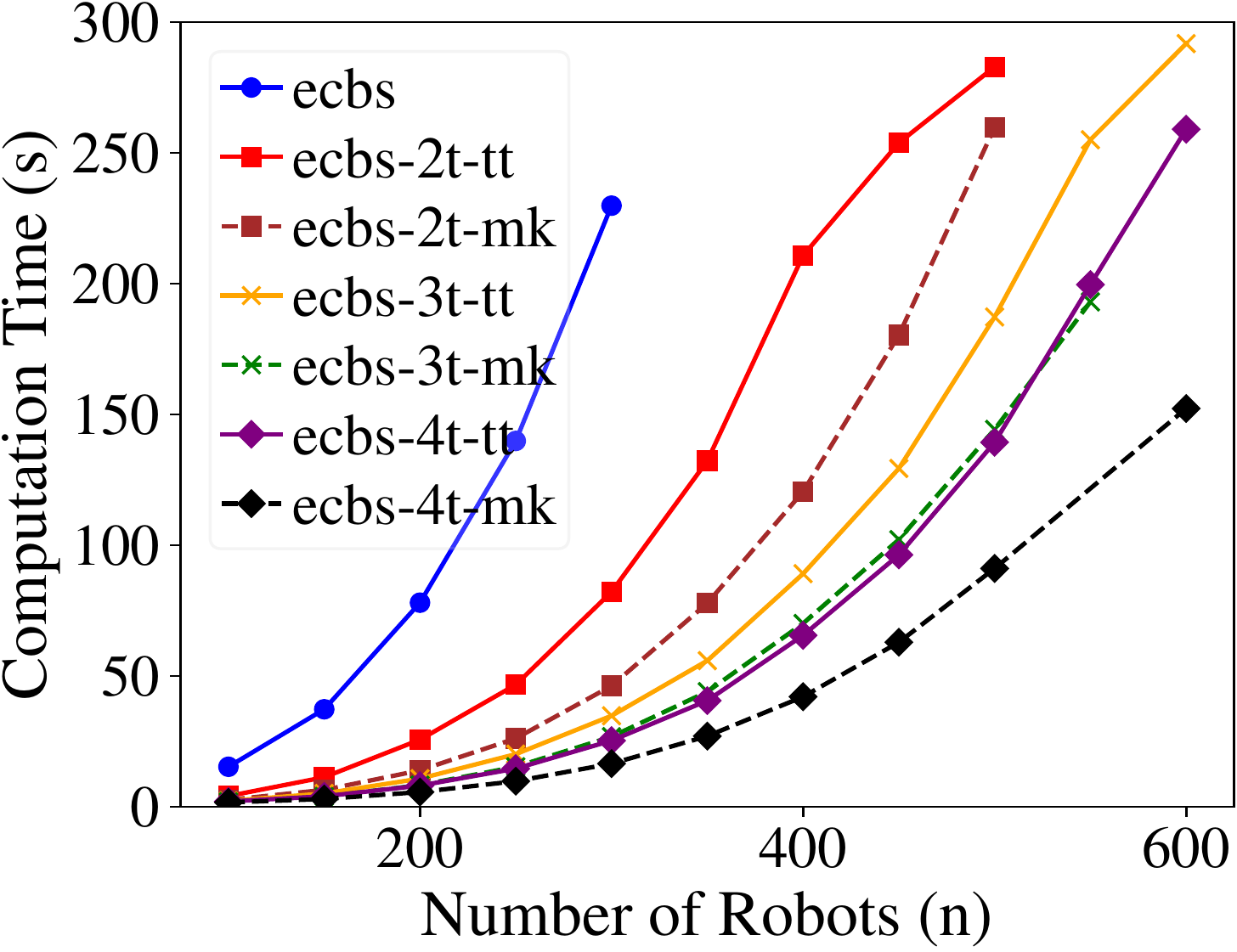}
    }
     \hfill
    \subfigure{
        \centering
        \includegraphics[width=0.465\linewidth]{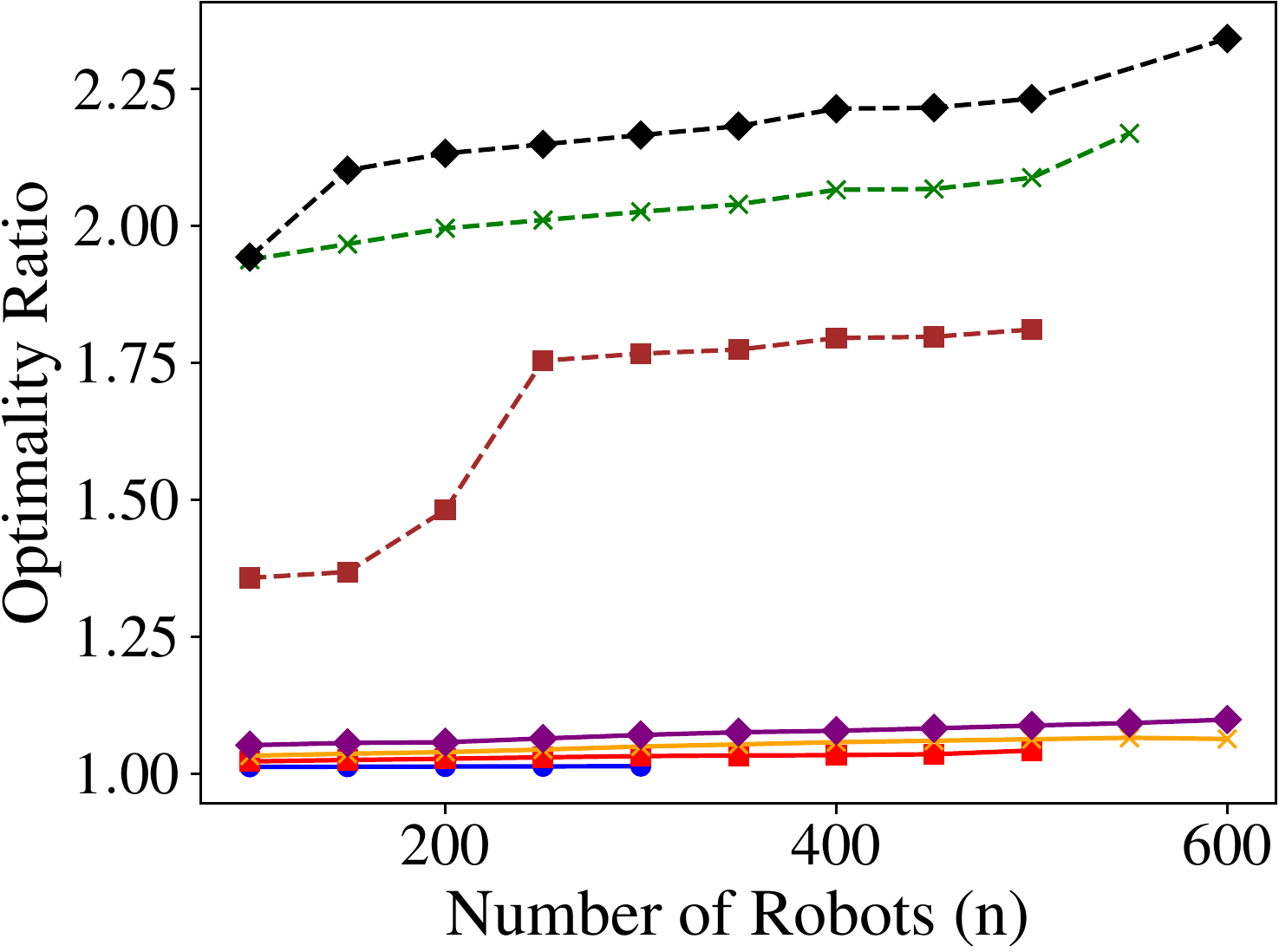}
   }
    \caption{\label{fig:time-ecbs-sum-of-costs-128} 
    Comparison of makespan and sum-of-costs time-split heuristics on {\em min-sum-of-costs} \mpp. The test graph is $128 \times 128$ grid with $10\%$ obstacles. }
\end{figure}

In the second scenario, we demonstrate that by modifying the way intermediate goals are generated, the time split heuristic becomes much better for the {\em min-sum-of-costs} objective (see Fig.~\ref{fig:time-ilp-sum-of-costs-32} and Fig.~\ref{fig:time-ecbs-sum-of-costs-128}). 
Here  "mk" stands for makespan and "tt" stands for sum of costs. A first observation is that, with the same number of sub-problems, the min-sum-of-costs version of time-split helps \ilp to generate solutions much closer to optimal, as compared to the original $k$-way split heuristic. For example, the optimality ratio dropped from $1.6$ to under $1.05$ when using the revised heuristic. We also find that, similar to the previous evaluation, the time-split heuristic reduces computation time. There is a small difference on the heuristic's effect on computation time when $k = 2$, since using min-sum-of-costs version of time-split heuristic makes the first sub-problem relatively harder than the second one. This implies that for sum-of-costs time-split, the threshold time span at the middle is not necessarily the optimal choice, which hints further opportunities for improvements.  

\begin{figure}[ht!]
  \centering
    \subfigure{
        \centering
        \includegraphics[width=0.29\linewidth]{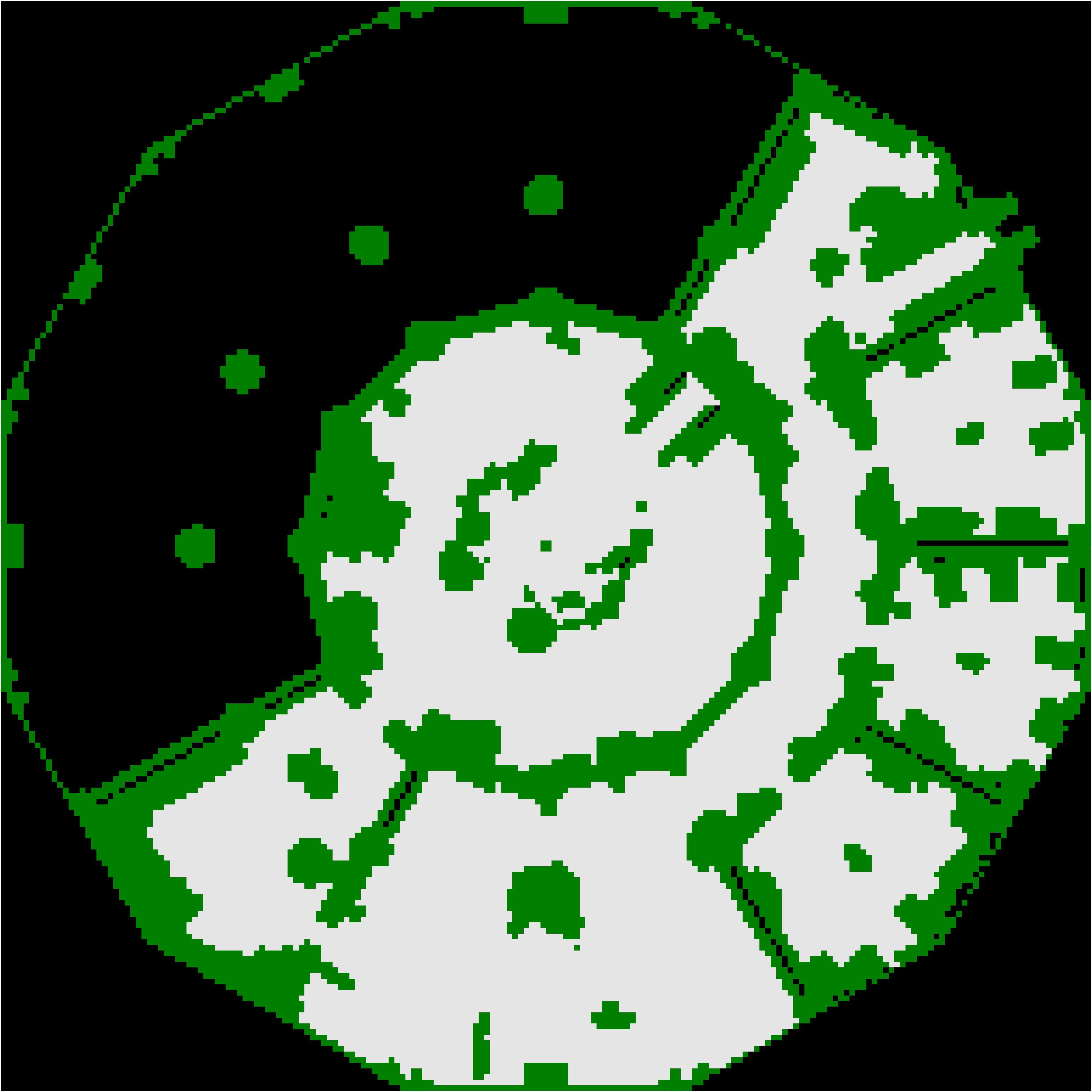}
  }
   \hfill
    \subfigure{
        \centering
        \includegraphics[width=0.29\linewidth]{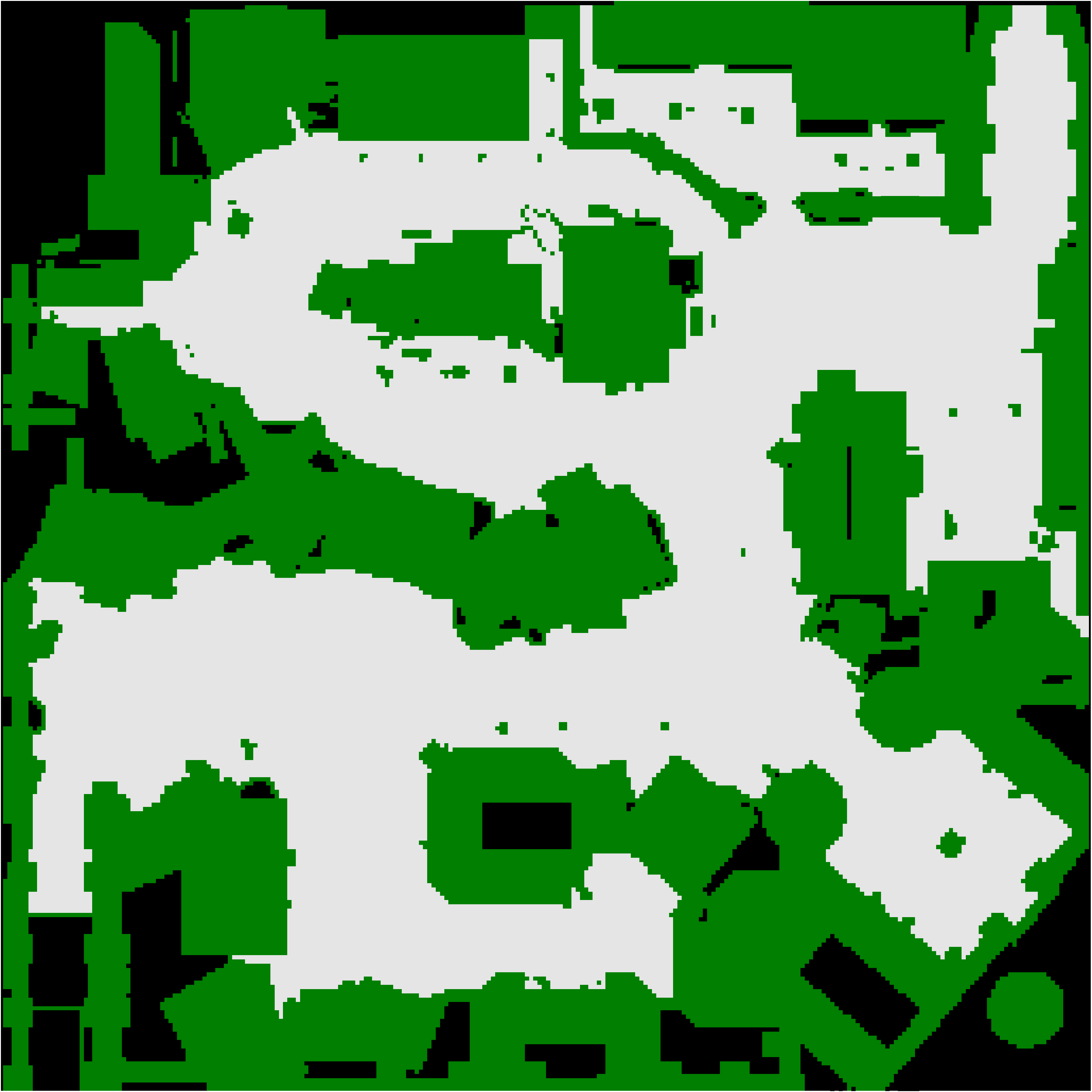}
    }
    \hfill
    \subfigure{
        \centering
        \includegraphics[width=0.29\linewidth]{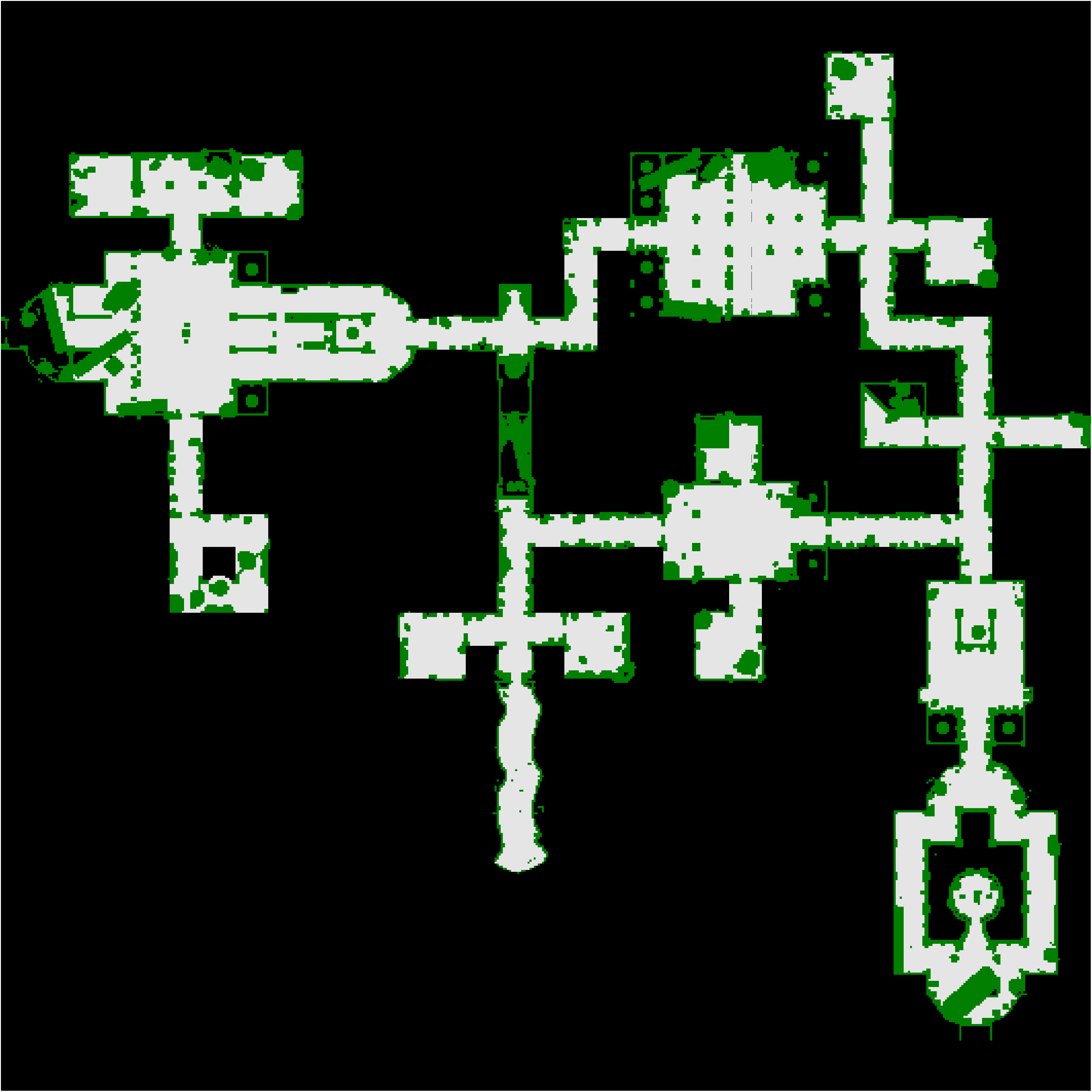}
    }
    \caption{\label{fig:DAO} The DAO maps: ost003d, den520d, brc202d. }
\end{figure}

\begin{figure}[ht!]
  \centering
    \subfigure{
        \centering
        \includegraphics[width=0.465\linewidth]{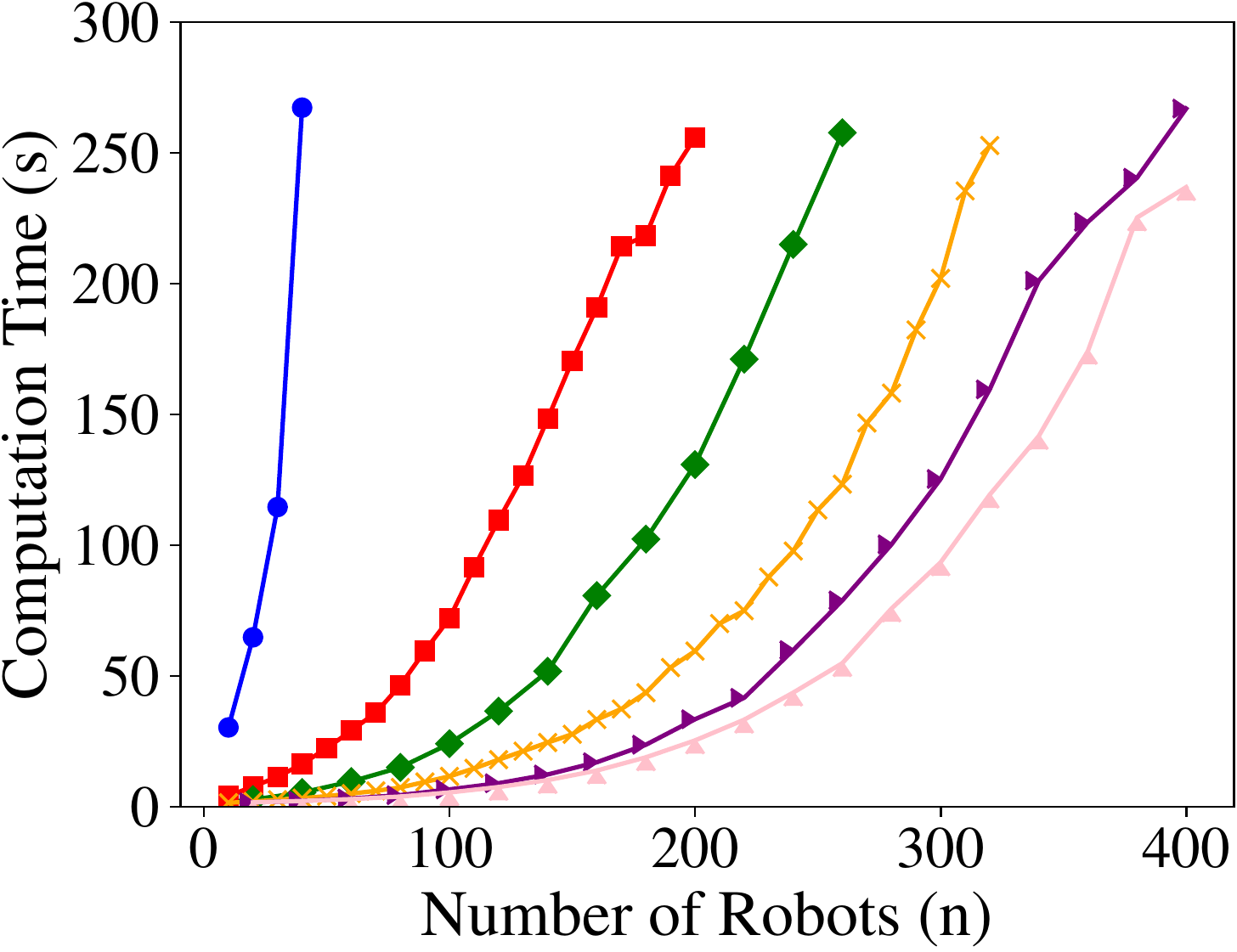}
  }
    \hfill
    \subfigure{
        \centering
        \includegraphics[width=0.465\linewidth]{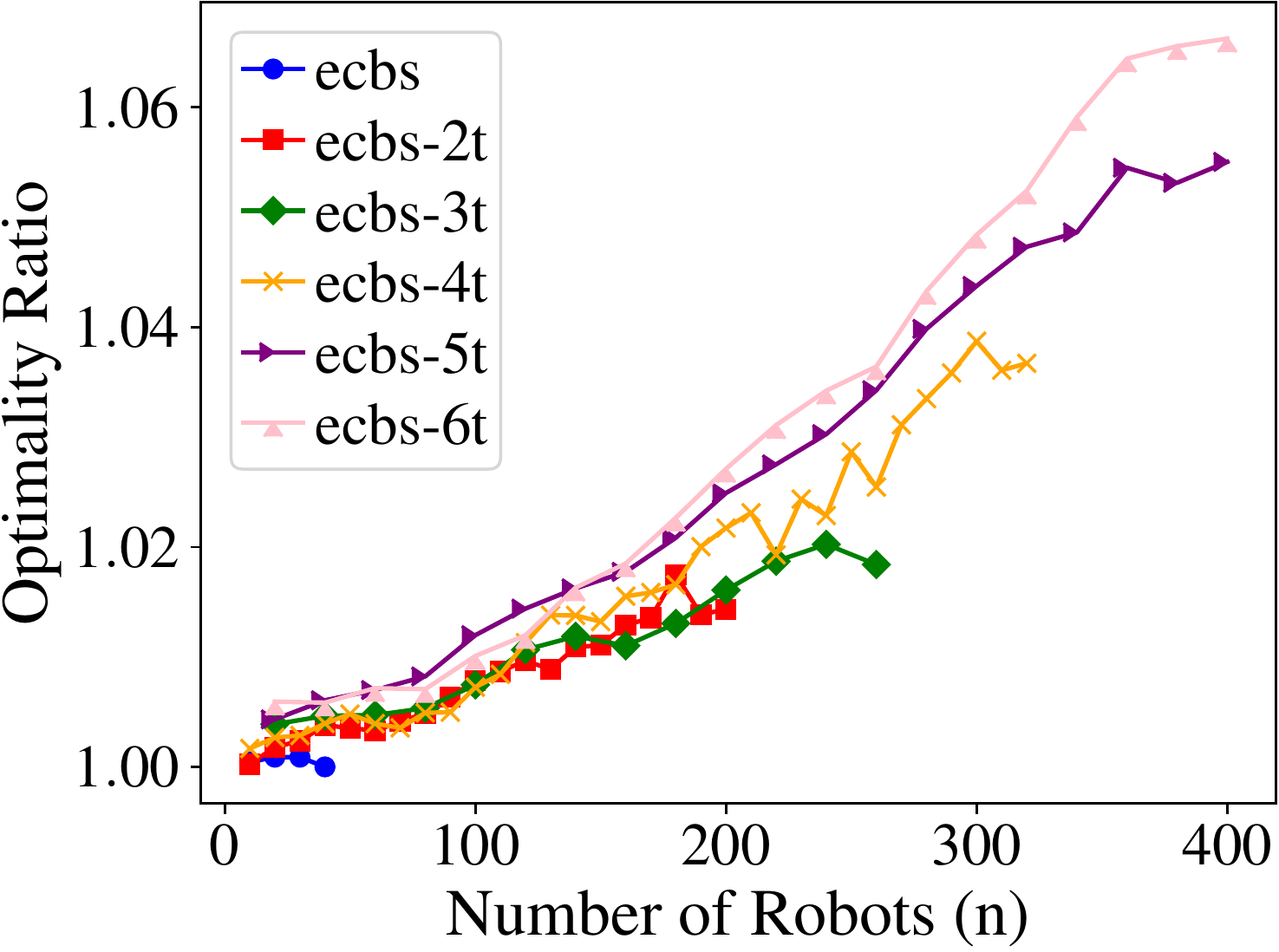}
    }
    \caption{\label{fig:time-ecbs-makespan-ost003d} Performance of time-split \ecbs on ost003d.  }
				\vspace{-2mm}
\end{figure}

\begin{figure}[ht!]
  \centering
    \subfigure{
        \centering
        \includegraphics[width=0.465\linewidth]{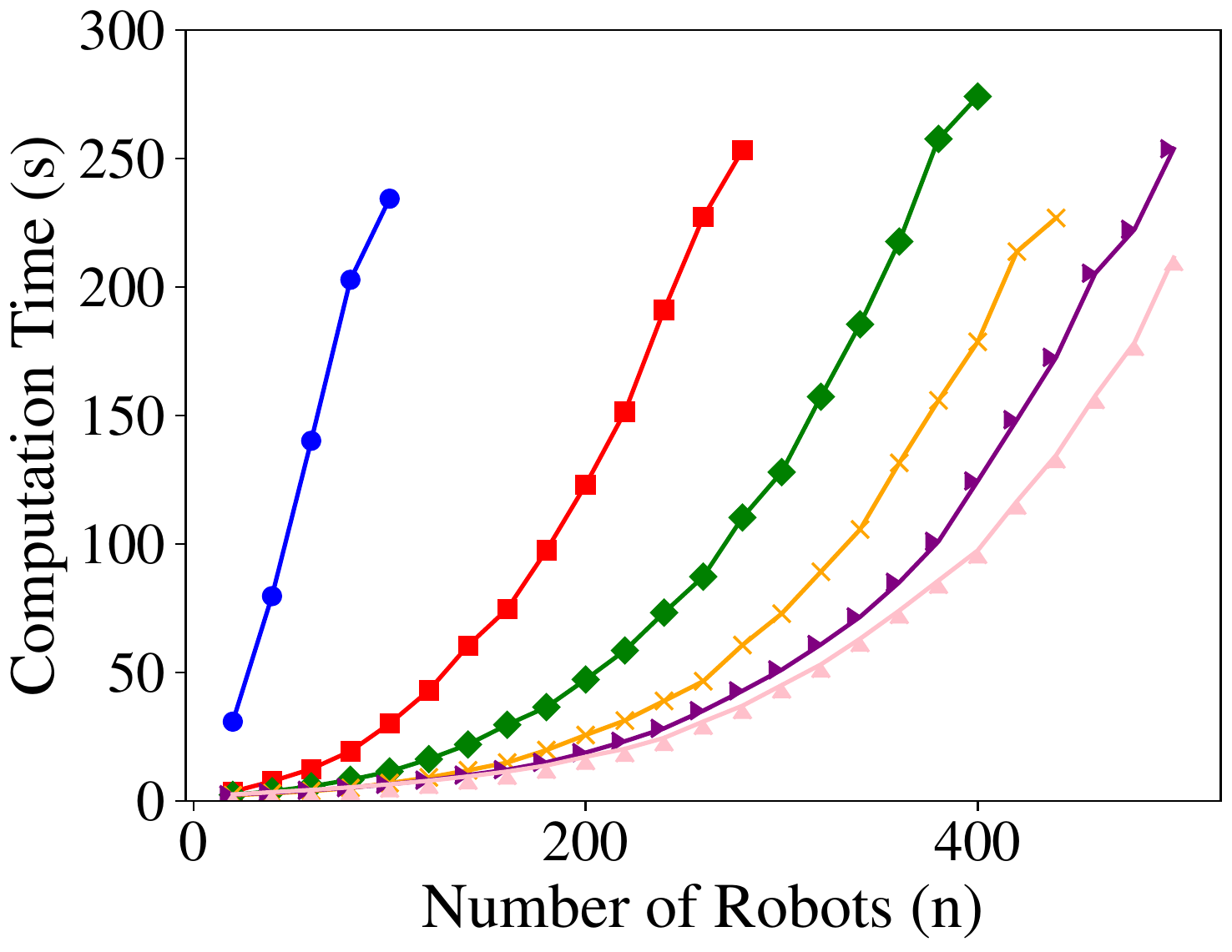}
   }
    \hfill
    \subfigure{
        \centering
        \includegraphics[width=0.465\linewidth]{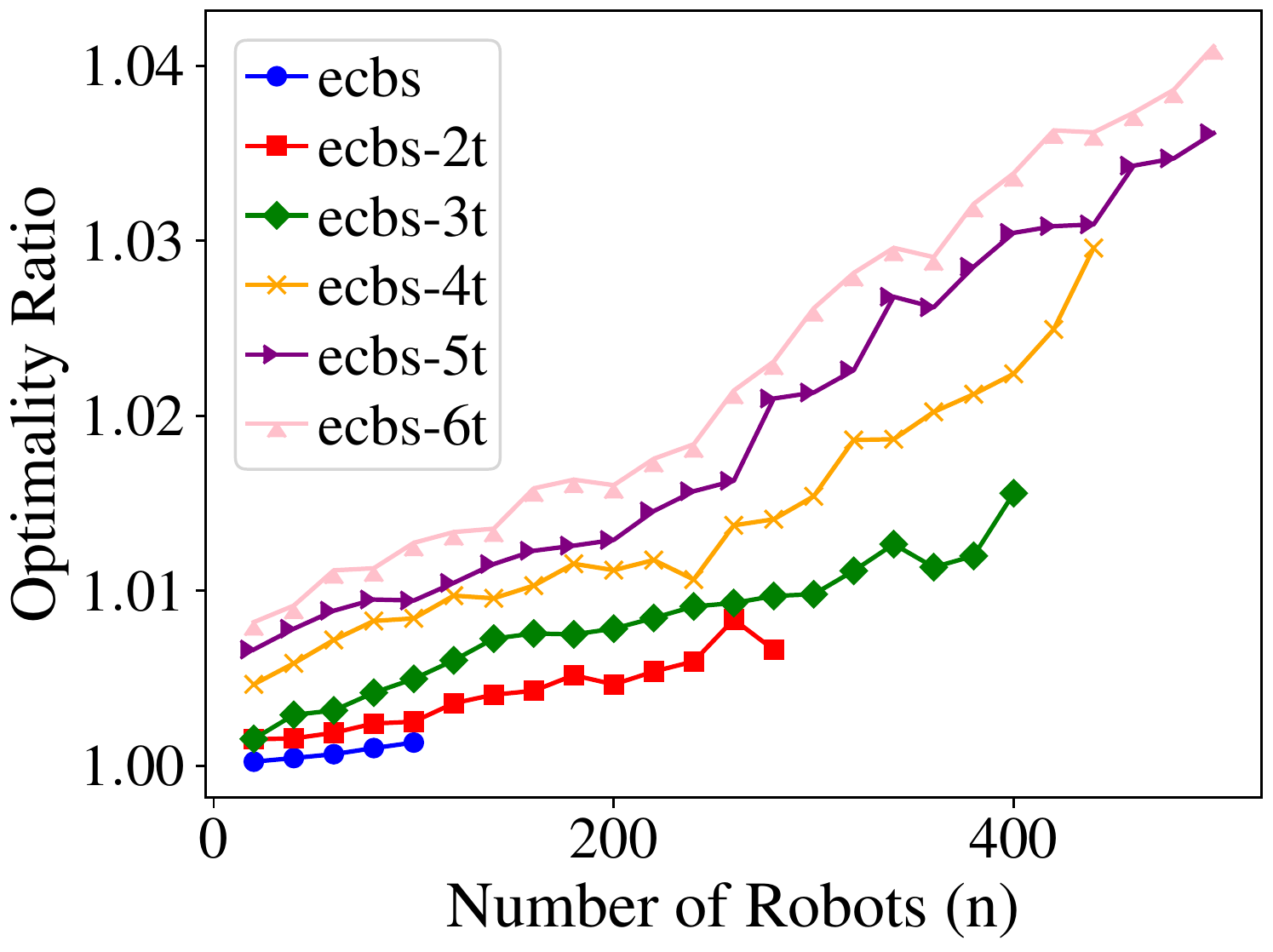}
    }
    \caption{\label{fig:den520d} Performance of time-split \ecbs on den520d.  }
				\vspace{-2mm}
  \end{figure}

\begin{figure}[ht!]
  \centering
    \subfigure{
        \centering
        \includegraphics[width=0.465\linewidth]{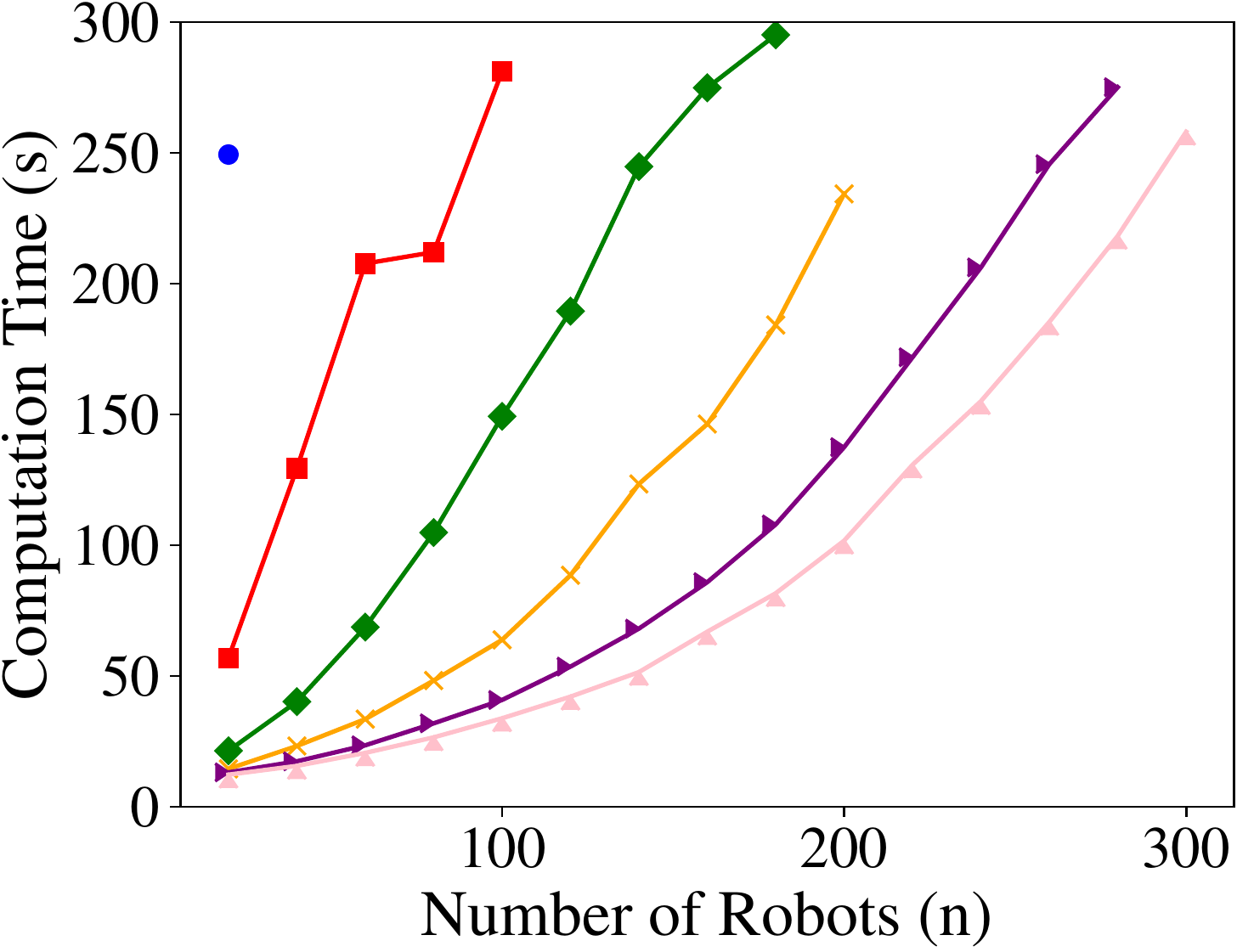}
    }
    \hfill
    \subfigure{
        \centering
        \includegraphics[width=0.465\linewidth]{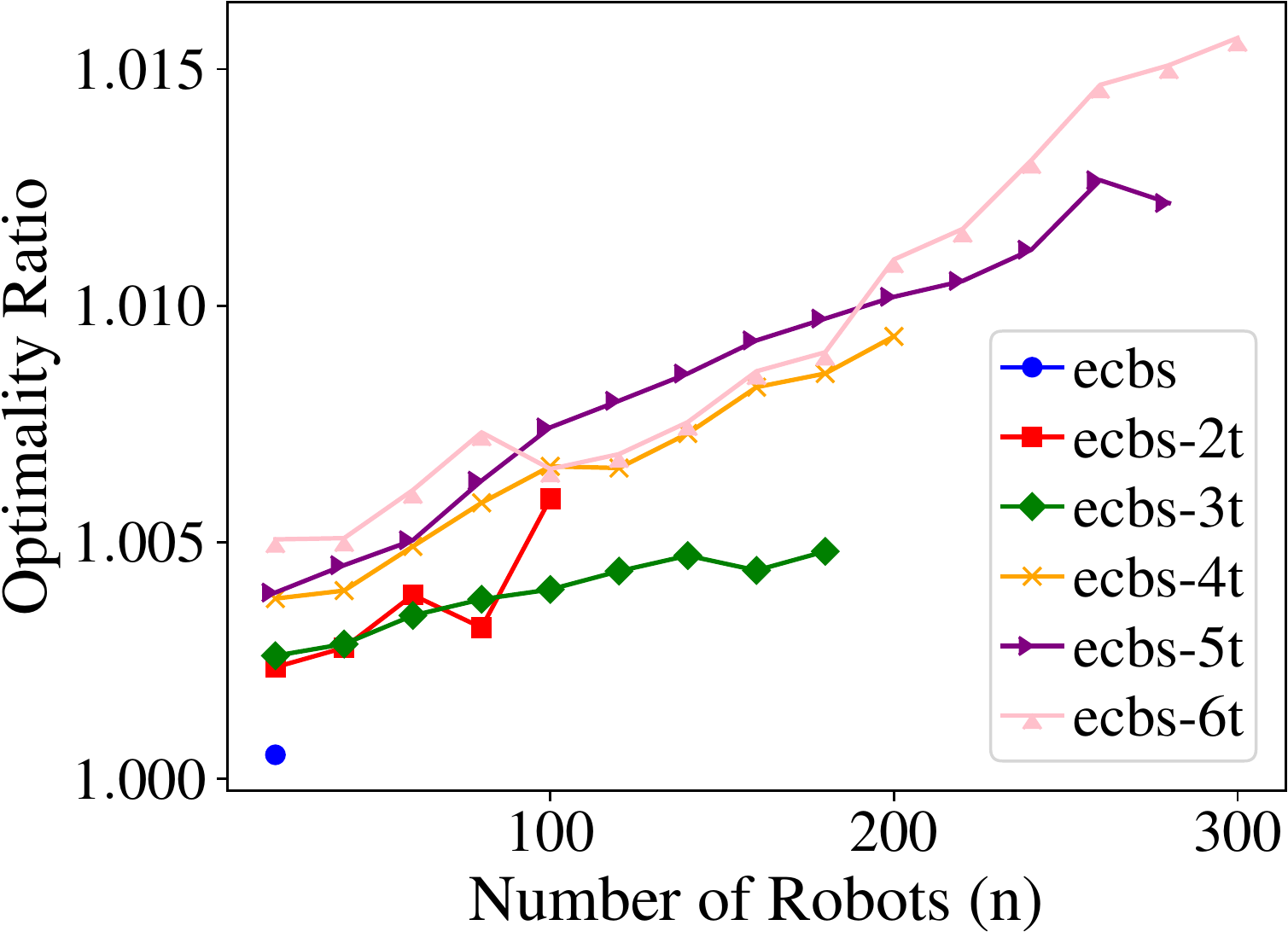}
    }
    \caption{\label{fig:brc202d} Performance of time-split \ecbs on brc202d.  }
				\vspace{-1mm}
  \end{figure}

Apart from the \ilp solver, the proposed time-split heuristic also applies to other solvers such as \ecbs. 
We evaluate the heuristic with \ecbs on both grid graphs and the Dragon Age Origins (DAO) maps~\cite{stern2019mapf}
(see Fig.~\ref{fig:DAO}), optimizing the makespan objective. 
Here, the test cases are imported from public \mpp benchmark instances that comes along with the maps, instead of randomly generated by ourselves. 
The results are  shown in Fig.~\ref{fig:time-ecbs-makespan-ost003d}-\ref{fig:brc202d}. Here, time-split allows problems with 10$\times$ more robots to be solved in the same amount of time while the solution optimality is just above 1.06. This further confirms that the time-split heuristic significantly extends the scalability of existing \mpp solvers while sacrificing little optimality. 

\subsection{Evaluation of the Space Split Heuristic}
\begin{figure}[ht!]
  \centering
   \subfigure{
        \centering
        \includegraphics[width=0.465\linewidth]{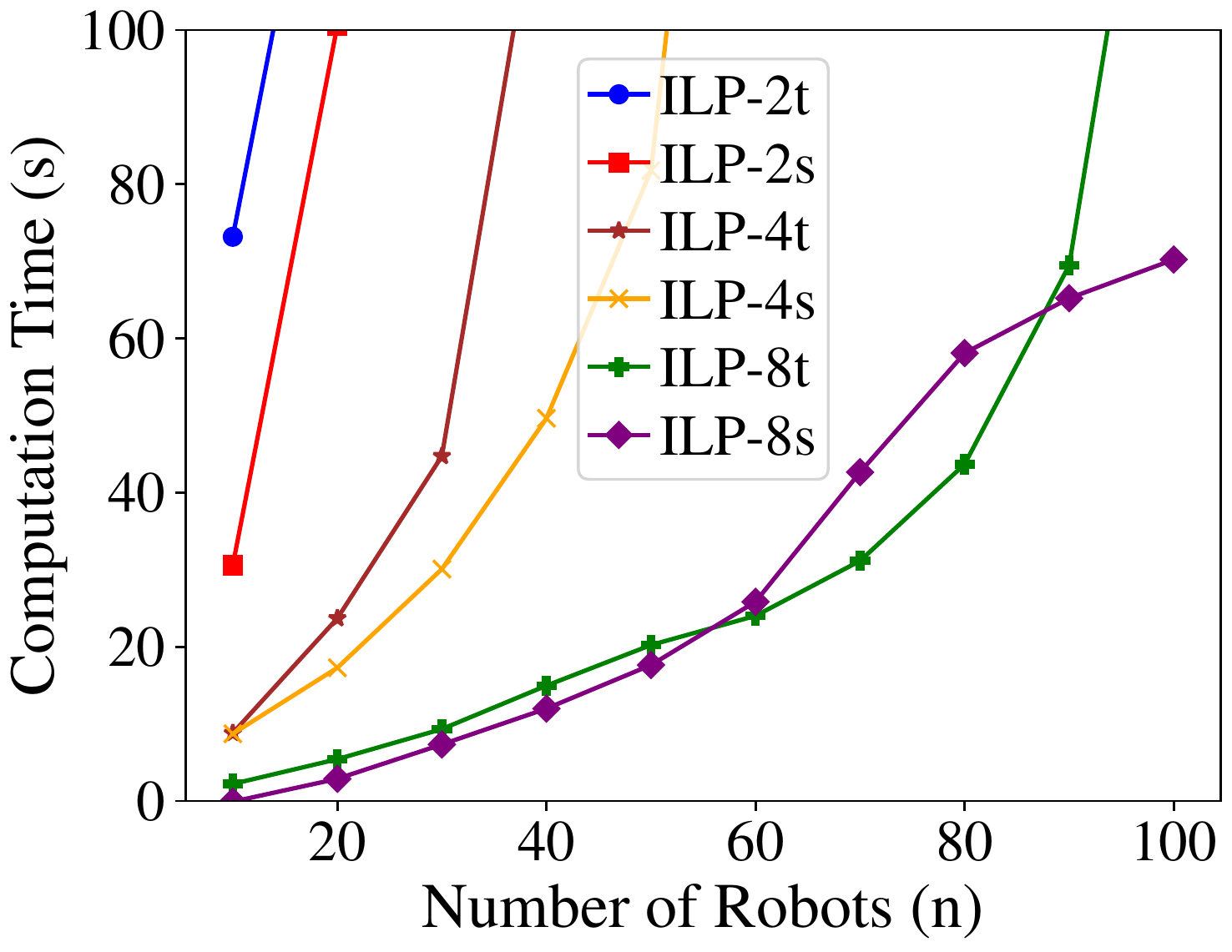}
    }
    \hfill
    \subfigure{
        \centering
        \includegraphics[width=0.465\linewidth]{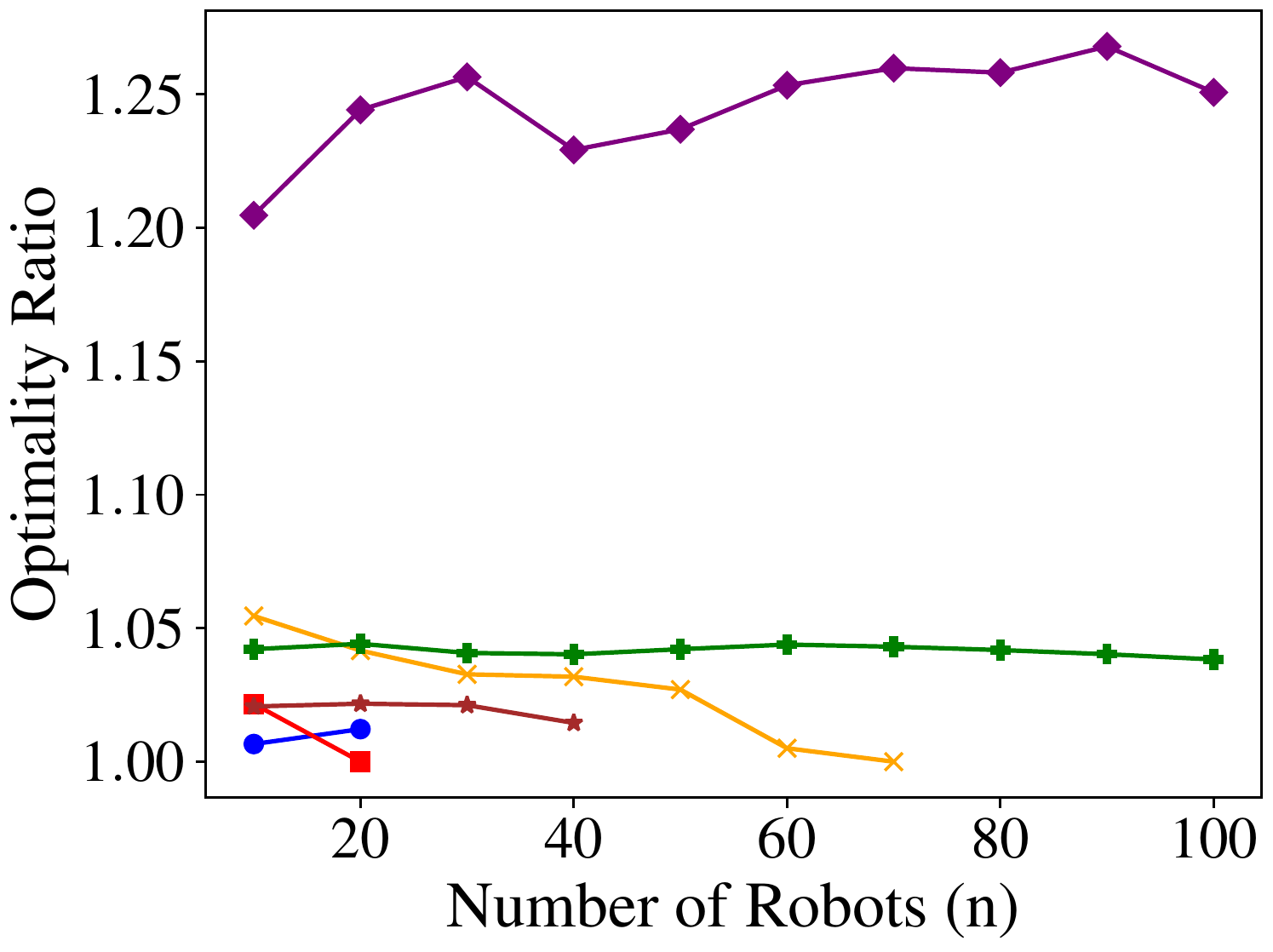}
    }
    \caption{\label{space} {\em{Time-split}} vs {\em{space-split}} \ilp on $128\times 128$ grid. }
\end{figure}

\vspace{-1mm}
For space split, we mainly focus on the makespan objective and evaluated over many types of large grid graphs.
In Fig.~\ref{space}, we test space-split \ilp on a $128\times 128$ grid and compare it with time-split. With the same split level $k$,  2-space and 4-space splits run faster than 2-time and 4-time splits, respectively. 4-time split runs out memory when there are 40 robots while 4-space split can handle 60-70 robots without out of memory error. Due to the fact that 8-space split needs to solve much more sub-problems but the number of CPUs is limited, 8-space split scales better but in some cases not faster than 8-time split.


\subsection{Combined time-space split heuristics}
\begin{figure}[ht!]
\centering
\subfigure{
    \centering
    \includegraphics[width=0.465\linewidth]{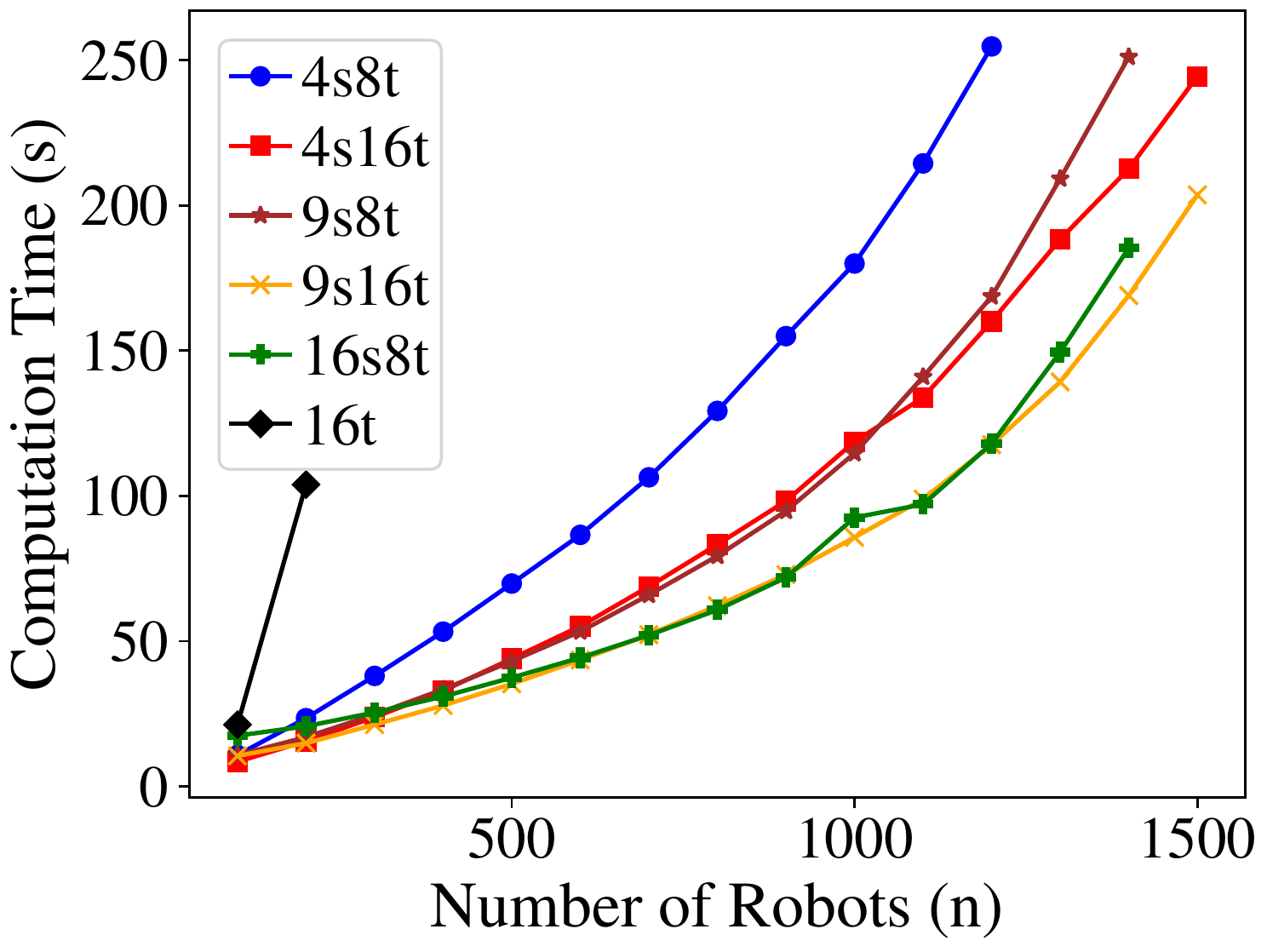}
}
\hfill
\subfigure{
    \centering
    \includegraphics[width=0.465\linewidth]{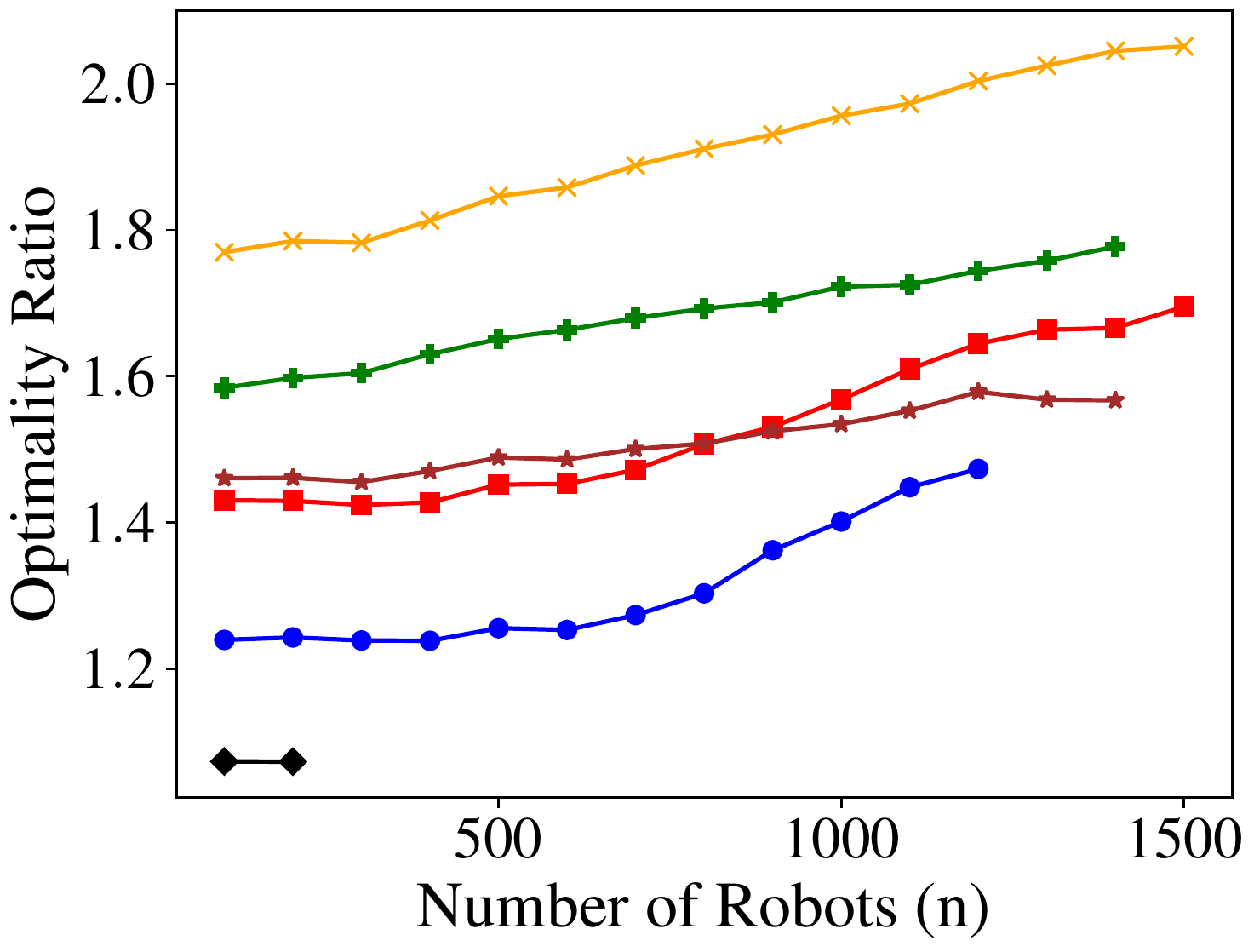}
}
\caption{\label{fig:time-space-split} Performance of time-space-split \ilp on a $128\times 128$ graph with $5\%$ obstacles.  }
\end{figure}

\vspace{-1mm}
As a last evaluation, we combine the two heuristics. Shown by the test result (Fig.~\ref{fig:time-space-split}) on a $128\times 128$ 
grid with $5\%$ obstacles, the combination of the two heuristics further extends 
existing algorithms' scalability. Method "xsyt" means that $y$-time-split is applied after a $x$-space split. While 16-time split has out of memory error when $n\geq 300$, time-space-split can handle instances with $n>1000$.
The \ilp solver can now be used to solve problems in large environments with a large number of robots, while maintaining $1.x$ optimality. 

\begin{figure}[ht!]
  \centering
   \subfigure{
        \centering
        \includegraphics[width=0.465\linewidth]{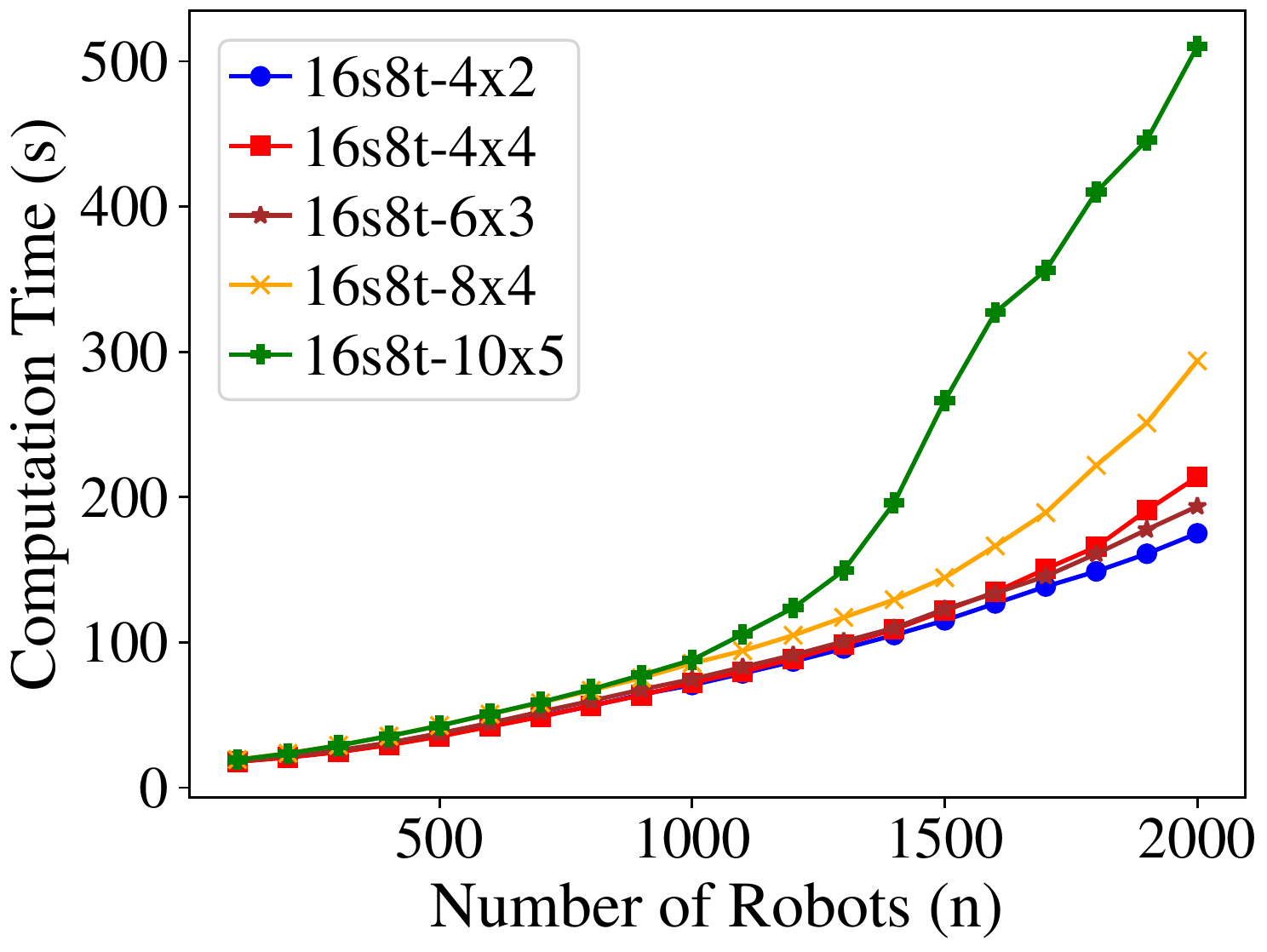}
    }
    \hfill
   \subfigure{
        \centering
        \includegraphics[width=0.465\linewidth]{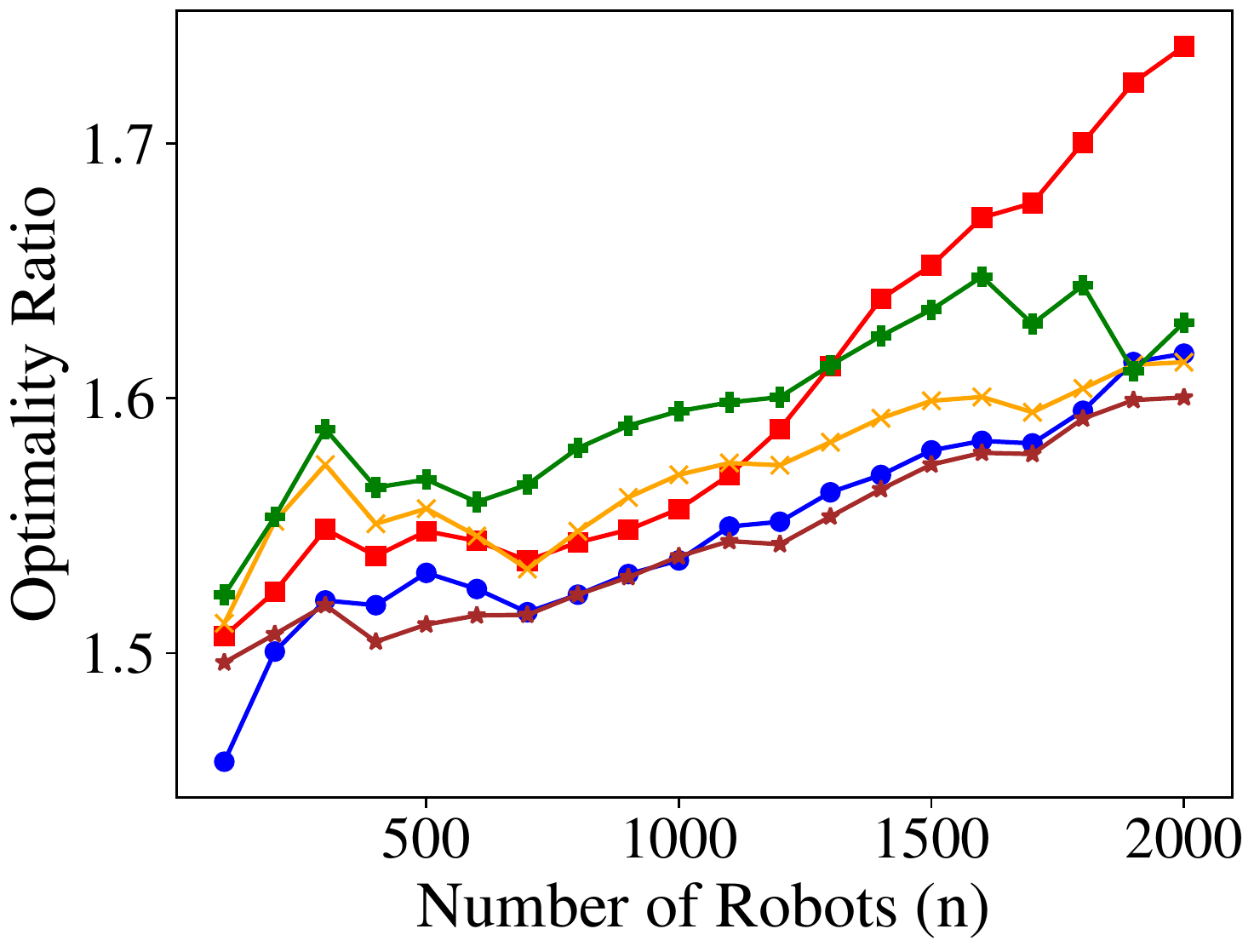}
    }
    \caption{\label{fig:difbuffer} Performance of space-split \ilp using different size of buffer zones on $128\times 128$ grid.  }
  \end{figure}

We also attempted different size of buffer zones to see how it would affect the performance of time-space-split (Fig.~\ref{fig:difbuffer}). As it shows, using smaller buffer zones (e.g., $4\times 2$) benefits both of computation time  and optimality.

\vspace*{-2mm}
\section{Conclusion}\label{sec:conclusion}
\vspace{-1mm}
In this work, temporal and spatial division heuristics are developed for improving the performance of \mpp solvers. These heuristics are shown to increase the computational speed while maintaining solution quality. 
These heuristics can be applied in combination with most \mpp algorithms. 
We note that (as proved) time split is complete and applicable to any graph; magnitudes of performance gains were consistently observed. On the other hand, space split is not complete. But space split enables \ilp to provide solutions of good quality for some challenging \mpp problems that are otherwise not solvable previously.

In future work, we intend to make these heuristics more {\em data-driven}. That is, we will determine how to perform the temporal and spatial division based on the problem input dynamically. For example, the $k$ in $k$-time-split can be selected based on $f(n,S,T)$ mentioned in the introduction. 
Furthermore, we plan to explore how to dynamically choose the buffer zone in the space split heuristic to improve its performance. Dynamic buffer zones are desirable when we work with irregular graphs and graphs with high obstacle density (e.g. $\geq 25\%$). Machine learning techniques may be also be applied. 

\bibliographystyle{IEEEtran}
\bibliography{all}

\clearpage

\end{document}